\pdfoutput=1
\documentclass[11pt]{article}

\usepackage[preprint]{acl}

\usepackage{times}
\usepackage{latexsym}

\usepackage[T1]{fontenc}
\usepackage[utf8]{inputenc}

\usepackage{microtype}

\usepackage{inconsolata}

\usepackage{graphicx}

\usepackage{amsmath}
\usepackage{amsthm}
\usepackage{booktabs}
\usepackage{algorithm}
\usepackage[switch]{lineno}
\usepackage{colortbl} 
\usepackage{xcolor}
\usepackage{amssymb}
\usepackage{algpseudocode}
\usepackage[normalem]{ulem}
\usepackage{multirow}
\useunder{\uline}{\ul}{}
\usepackage{tcolorbox}
\usepackage{placeins}

\title{Mitigating Hallucinations in Large Vision-Language Models by Self-Injecting Hallucinations}

\author{
\textbf{Yifan Lu\textsuperscript{1,2,3}},
\textbf{Ziqi Zhang\textsuperscript{1,2}},
\textbf{Chunfeng Yuan\textsuperscript{1,2,3}\thanks{Corresponding author.}},
\textbf{Jun Gao\textsuperscript{4}},
\textbf{Congxuan Zhang\textsuperscript{5}}, \\
\textbf{Xiaojuan Qi\textsuperscript{6}},
\textbf{Bing Li\textsuperscript{1,2,3}},
\textbf{Weiming Hu\textsuperscript{1,2,3,7}},
\\
\textsuperscript{1}Beijing Key Laboratory of Super Intelligent Security of Multi-Modal 
Information, CASIA \\
\textsuperscript{2}State Key Laboratory of Multimodal Artificial Intelligence Systems, CASIA\\
\textsuperscript{3}School of Artificial Intelligence, University of Chinese Academy of Sciences\\
\textsuperscript{4}Hello Group
\textsuperscript{5}Nanchang Hangkong University
\textsuperscript{6}The University of Hong Kong \\
\textsuperscript{7}School of Information Science and Technology, ShanghaiTech University \\
\texttt{luyifan2021@ia.ac.cn,cfyuan@nlpr.ia.ac.cn} 
}

\begin{document}
\maketitle

\begin{abstract}
Large Vision-Language Models (LVLMs) suffer from serious hallucination problems, where the model-generated responses are inconsistent with the visual inputs. 
Existing hallucination mitigation methods are mainly based on preference alignment and require external human annotations or auxiliary models for preference data collection, which increase costs and limit sustainable improvement.
To tackle these challenges, we propose \textbf{A}utonomous \textbf{P}reference \textbf{A}lignment via \textbf{S}elf-\textbf{I}njection (\textbf{APASI}), a novel and generalizable method that mitigates hallucinations without external dependencies. APASI leverages the target LVLM to self-inject hallucinations into a generated response, creating a pair of responses with varying preference levels. 
During the self-injection process, the dis-preferred response is generated based on three key observations of hallucinations, ensuring it simulates real hallucination patterns. This fidelity offers an accurate learning signal for hallucination mitigation.
Moreover, APASI incorporates an iterative alignment training strategy combined with curriculum learning to periodically update the preference data with increasing challenge, enabling stable and continuous enhancement of the LVLM.
Extensive experiments across six benchmarks show that APASI not only effectively mitigates hallucinations for three baseline models but also achieves comparable or even superior performance to alignment-based methods with external dependency, thereby demonstrating its effectiveness and generalization capability. The code is available at \href{https://github.com/davidluciolu/APASI}{https://github.com/davidluciolu/APASI}.
\end{abstract}

\section{Introduction}

% LVLM and hallucination problem
The study of Large Vision-Language Models (LVLMs) 
% \cite{zhu2023minigpt,li2023mimic,liu2024improved,liu2024llava,wang2024qwen2}
\cite{li2023mimic,liu2024improved,liu2024llava,wang2024qwen2}
has made remarkable progress in recent years. 
% LVLMs integrate visual perception with the powerful language reasoning and generative capabilities of Large Language Models (LLMs)
% % \cite{achiam2023gpt,chiang2023vicuna,touvron2023llama}
% \cite{touvron2023llama,chiang2023vicuna}
% , thereby substantially enhancing cross-modal understanding in vision-language tasks and achieving notable performance across various applications. 
LVLMs substantially enhance cross-modal understanding in vision-language tasks and achieve notable performance across various applications. 
Despite their demonstrated efficacy, LVLMs frequently encounter hallucination problems which refer to the inconsistency between the factual content of visual input and the corresponding generated textual response \cite{liu2024survey, yan2024evaluating}. 
% The hallucinated response typically includes visual elements that are not depicted in the image.
This problem undermines the reliability of LVLMs, making the mitigation of hallucinations a critical area of research.

\begin{figure}[t]
	\begin{center}
		%\fbox{\rule{0pt}{2in} \rule{.9\linewidth}{0pt}}
		\includegraphics[width=\linewidth]{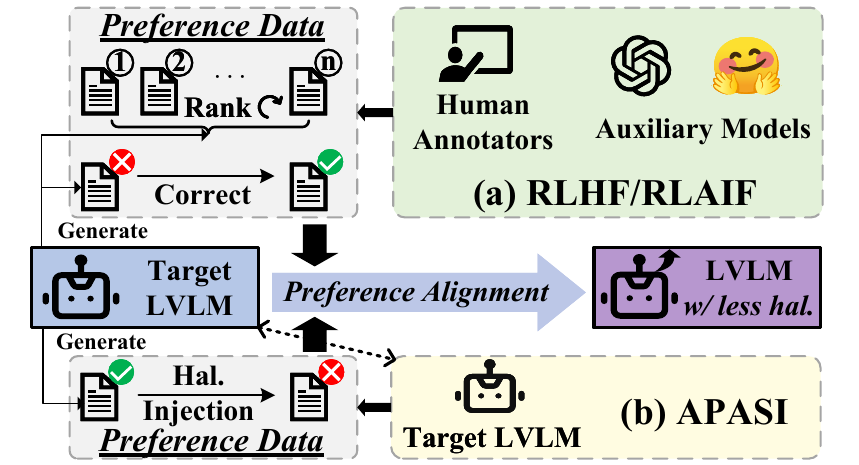}
	\end{center}
	\caption{Comparison of different alignment-based hallucination mitigation methods. While all methods employ the framework of preference alignment to optimize models using preference data, they vary in their approaches of collecting the data. Best viewed in color.}
	\label{fig:overview}
\end{figure}

% current alignment based methods 
Various methods have been proposed to mitigate hallucinations in LVLMs through preference alignment techniques, such as Reinforcement Learning from Human Feedback (RLHF) \cite{sun2023aligning,yu2024rlhf} and Reinforcement Learning from AI Feedback (RLAIF) \cite{zhao2023beyond,yu2024rlaif}.
As shown in Fig.\ref{fig:overview}(a), the target LVLM is trained using ranking or corrective preference data to align with the preference for {\em no hallucinations} or {\em reduced hallucinations}. 
% drawbacks of RLAIF
Notably, RLAIF-based methods alleviate the cost and subjectivity issues in RLHF preference data collection \cite{li2023silkie} by replacing human annotators with auxiliary feedback models such as GPT-4 \cite{achiam2023gpt}. Methods \cite{zhou2024calibrated,ouali2025clip} further promote these improvements by replacing proprietary models with open-source models. 
% However, these methods fundamentally rely on the capabilities of auxiliary feedback models, which impose a limit on the sustainable improvement of the target LVLM.
% As alignment progresses, the capabilities of the target LVLM become comparable to those of the feedback models \cite{yu2024rlaif}, particularly in handling hallucinations. At this point, the feedback models may fail to detect minor hallucinations in the LVLM’s responses or provide effective corrections, leading to optimization saturated.
However, as alignment progresses, the capabilities of the target LVLM become comparable to the fixed and finite capabilities of the feedback models \cite{yu2024rlaif}, particularly in handling hallucinations. 
At this point, the feedback models may struggle to detect minor hallucinations in the LVLM’s responses or provide effective corrections.
% thereby imposing a limit on the sustainable improvement of the target LVLM.
Therefore, methods that rely on external models face limitations in achieving sustainable improvement of the target LVLM.

% APASI operates independently of external models or annotations, minimizing the data cost and reducing reliance on external resources.

% respond to existing limitations: self-improment style

% proposed method overview
To overcome these limitations, we propose \textbf{A}utonomous \textbf{P}reference \textbf{A}lignment via \textbf{S}elf-\textbf{I}njection (\textbf{APASI}), a novel hallucination mitigation method that constructs preference data using the target model itself.
Though following the common framework of using the Direct Preference Optimization (DPO) \cite{rafailov2024direct} algorithm to iteratively align the target LVLM with the preference for {\em reduced hallucinations}, APASI distinctively utilizes self-generated data, as shown in Fig.\ref{fig:overview}(b).
% APASI is designed to autonomously generate valid preference pairs for DPO, based on the idea that injecting hallucinations into responses is simpler and more feasible than collecting ranking or corrective feedback as typically done in RLHF/RLAIF methods. 
% Specifically, in each preference pair, the preferred response is a model-generated response to image-text input, which is then injected with hallucinations LVLM with the visual encoder disabled, resulting in the dis-preferred response.
Instead of collecting ranking or corrective feedback as typically done in RLHF/RLAIF methods, APASI generates its data by injecting hallucinations into model-generated responses, creating effective dis-preferred responses that simulate real hallucination patterns. With the original model-generated responses as the preferred, this process forms valid preference pairs for DPO. 
Meanwhile, the straightforward injection of hallucinations is autonomously executed independently of external auxiliary models, facilitating sustainable model improvement.
Moreover, APASI is easily scalable as it requires no annotations.

% In this section, we introduce APASI, an autonomous preference alignment method designed to mitigate hallucinations in LVLMs. As illustrated in Fig.\ref{fig:overview},  APASI autonomously generates preference pairs $(y^+, y^-)$ required by DPO from unannotated data. Given an image and a text prompt, APASI first employs a baseline LVLM to generate response as the preferred response $y^+$. By disabling the visual encoder from the baseline LVLM, APASI creates a language-only {\em blind} LVLM that injects hallucinations into hallucinations into $y^+$ to produce a corrupted response as the dis-preferred $y^-$. This process forms the preference pair $(y^+, y^-)$, which is subsequently utilized for DPO tuning, alongside the original input image and instruction. APASI also applies an iterative alignment strategy with curriculum learning to update preference data with increasing difficulty during the training process.
% The entire APASI pipeline operates independently of any annotations or external teacher models. 
% Further details about APASI are discussed in the following subsections.

% \begin{figure}[t]
% 	\begin{center}
% 		%\fbox{\rule{0pt}{2in} \rule{.9\linewidth}{0pt}}
% 		\includegraphics[width=\linewidth]{overview1.pdf}
% 	\end{center}
% 	\caption{Overview of the proposed APASI. APASI constructs preference pairs from unannotated data by self-injecting hallucinations into original model responses. Best viewed in color.}
% 	\label{fig:overview}
% \end{figure}

% enhancement for self-injection
% targeted to real hallucination patters
However, directly injecting hallucinations into responses through bad prompts or visual corruption, as done in \cite{zhou2024aligning,deng2024enhancing}, fails to accurately simulate real hallucination patterns. 
The self-injection process in APASI is based on three key observations of hallucinations:
1) LVLMs are prone to hallucinate objects that frequently \textbf{co-occur} with the existent objects in the image \cite{li2023evaluating};
2) LVLMs are prone to generate hallucinated content with an over-reliance on \textbf{language priors} \cite{favero2024multi};
3) Hallucinations in LVLMs typically cluster towards the \textbf{latter part} of the response \cite{zhou2024analyzing}. 
Accordingly, the target LVLM is guided to fabricate sentences about non-existent \textbf{co-occurring} objects with \textbf{language-only} inputs and then integrates this hallucinated content into the \textbf{latter parts} of the preferred response, forming the dis-preferred response.
In this way, the preference pair facilitates an accurate learning signal for the mitigation of real hallucination patterns. 
Moreover, as learning progresses, APASI incorporates a curriculum \cite{bengio2009curriculum} to gradually reduce the injection of hallucinations, making it more challenging to distinguish subtler differences in the preference pairs, thereby helping to refine the LVLM’s ability to identify hallucinations smoothly over iterations.

% iterative learning and curriculum, emphasis cl instead of ia

% For training, APASI employs an iterative alignment strategy that cycles between data collection and model optimization to alleviate the distribution shift problem \cite{gao2023scaling,yu2024rlaif} and to promote continuous improvement.
% Moreover, to facilitate a more effective learning trajectory, the iterative strategy incorporates curriculum learning \cite{bengio2009curriculum}, which updates the preference data to be more challenging by increasing the hallucination injection rate over iterations. This tailored curriculum helps refine the model’s ability to distinguish the preference.
% using the latest optimized model
% For training, APASI employs an iterative alignment scheme with curriculum learning \cite{bengio2009curriculum} to promote continuous improvement. As alignment progresses, the preference data is updated to be more challenging than those in the last iteration. This tailored curriculum fosters a gradual and smooth refinement of the LVLM’s ability to distinguish hallucinations  

Our contributions are summarized as follows:
\begin{itemize}
\item We propose APASI, a novel hallucination mitigation method for LVLMs. APASI designs a scalable and effective pipeline to autonomously collect preference data by self-injecting hallucinations into model-generated responses, thereby minimizing reliance on external data sources and enabling sustainable improvement.

\item APASI leverages key insights into hallucination patterns to accurately construct preference pairs, providing a precise learning signal for hallucination mitigation.
APASI further incorporates an iterative alignment strategy with curriculum learning for stable improvement.

\item Extensive experiments on various benchmarks validate that APASI effectively mitigates the hallucination problem and enhances performance for baselines including LLaVA-v1.5, LLaVA-v1.6, and Qwen2-VL, showcasing its efficacy and generalization capability.

\end{itemize}

\section{Related Works}
\subsection{Hallucination in LVLMs}
% In recent years, significant progress has been made in the study of Large Vision-Language Models (LVLMs) \cite{zhu2023minigpt,li2023mimic,liu2024improved,liu2024llava,wang2024qwen2}, which are built by connecting visual encoders \cite{radford2021learning} with powerful Large Vision-Language Models (LLMs) \cite{achiam2023gpt,chiang2023vicuna,touvron2023llama} with strong reasoning and generative capabilities.
% LVLMs have achieved impressive performance in various vision-language tasks, leveraging the capabilities of LLMs and the instruction tuning paradigm \cite{sanh2022multitask} that unifies various tasks into a single response generation to vision-language input. 

% \subsection{Hallucination in LVLMs}
Despite the success, current LVLMs suffer from hallucination problems, where the generated response is inconsistent with the visual inputs. Specifically, this inconsistency is multi-facet, including errors in object, attribute, and relationship \cite{liu2024survey}.
% The misalignment of language and vision modalities is a key factor in hallucination \cite{sun2023aligning,zhao2023beyond}. 
Recent research finds key observations of hallucinations including: \textbf{Object co-occurrence} \cite{li2023evaluating,zhou2024analyzing,leng2024mitigating}, \textbf{Language prior} \cite{leng2024mitigating,favero2024multi}, and \textbf{Positional factor} \cite{zhou2024analyzing,favero2024multi}.
% 1) \textbf{Object co-occurrence} \cite{li2023evaluating,zhou2024analyzing,leng2024mitigating}: LVLMs are prone to hallucinate objects that frequently co-occur with the existent objects in the image.
% 2) \textbf{Language prior} \cite{leng2024mitigating,favero2024multi}: LVLMs are prone to generate hallucinated content with over-reliance on language priors.
% 3) \textbf{Positional factor} \cite{zhou2024analyzing,favero2024multi}: Hallucinations tend to appear more frequently in the latter parts of the response.
Our proposed APASI is designed to focus on these key observations to construct valid preference pairs.

\subsection{Hallucination Mitigation via Alignment}
% To mitigate hallucinations in LVLMs, various strategies have been proposed. 
% Early researches include constructing high-quality datasets for instruction tuning \cite{liu2023mitigating}, 
% detecting and correcting hallucinations post-generation \cite{zhou2024analyzing,yin2024woodpecker,wu2024logical,yue2024less}, 
% and optimizing the model's decoding process to enforce appropriate attention patterns during generation\cite{favero2024multi,leng2024mitigating,huang2024opera}.
% Though these strategies alleviate hallucination to some extent, they do not focus directly on
% improving modality alignment.
% Specifically, object hallucination is a focus for recent research and occurs when the responses erroneously including objects that are not present the visual inputs.
Preference alignment \cite{ji2023ai} has become a prominent strategy for mitigating hallucinations in models, aiming to align model behavior with the human preference for {\em no hallucinations} or {\em reduced hallucinations} \cite{zhao2023beyond,yu2024rlhf}. 
Early methods \cite{yu2024rlhf} use RLHF to collect preference data from human annotators and utilize optimization algorithms such as Proximal Policy Optimization \cite{schulman2017proximal} and Direct Preference Optimization \cite{rafailov2024direct} to fine-tune the model. 
The preference data are typically gathered either by direct collection of rankings for sampled responses \cite{sun2023aligning} or by correcting hallucinations in model-generated responses to establish a preferred version \cite{gunjal2024detecting}.
Some studies \cite{li2023silkie,zhao2023beyond,zhou2024aligning} use RLAIF to replace human annotators with powerful proprietary models such as GPT-4 \cite{achiam2023gpt}, reducing costs and enhancing annotation quality.
Methods \cite{zhou2024calibrated,yu2024rlaif,ouali2025clip} further reduce the cost by using open-source models such as CLIP \cite{radford2021learning} and LLaVA-v1.6 \cite{liu2024llava}.

% \subsection{Hallucination Mitigation via Self-Improvement}
Recent studies have explored self-improvement mechanisms \cite{huang2023large,chen2024self}, where preference data are derived from the optimized model itself. 
% SIMA \cite{wang2024enhancing} prompts the currently optimized LVLM with carefully designed critic prompt to rank multiple sampled responses and form preference pairs. However, the self-critic process requires ground-truth responses, limiting its scalability to unlabeled data. 
SIMA \cite{wang2024enhancing} uses critic prompts to self-rank sampled responses, but relies on ground-truth references, limiting scalability.
STIC \cite{deng2024enhancing} generates dis-preferred responses through 
misleading questions or corrupted images, eliminating the need for ground truth. 
% Notably, our proposed APASI creates dis-preferred responses by directly injecting hallucinations into original responses based on the key observations of hallucinations. Furthermore, APASI is free from external model, annotations, and complicated prompt design.
In contrast, APASI directly injects hallucinations based on key hallucination patterns, requiring no external models, annotations, or complex prompt design.

\section{Methodology}

We first provide preliminaries of the DPO algorithm in Section\ref{ss:3.1}. 
Then in Section\ref{ss:3.2}, we dive into the preference data construction process based on pertinent self-injection of hallucination to provide an accurate learning signal for DPO training.
Finally, we introduce the training scheme for the APASI framework using iterative alignment strategy with curriculum learning to achieve continuous and stable optimization in Section\ref{ss:3.3}.

\subsection{Direct Preference Optimization}
\label{ss:3.1}
The LVLM with parameters $\theta$ denoted as $M_\theta$, defines a conditional distribution $p_\theta(y|v, x)$, where
$y$ denotes the output response for the input image $v$ and the text prompt $x$.
% The proposed APASI leverages DPO \cite{rafailov2024direct}, a widely used preference learning algorithm, to tune the parameters $\theta$ and align the LVLM toward the preference of reduced hallucinations. DPO learns from the preference data defined as $\mathcal{D} = \{(v_i, x_i, y^+_i, y^-_i)\}_{i=1}^N$, where the preference pair $(y^+_i, y^-_i)$ consists of a preferred and a dis-preferred response for the input $v_i$ and $x_i$. 
The proposed APASI leverages DPO \cite{rafailov2024direct} to tune the parameters $\theta$ and align the LVLM toward the preference of reduced hallucinations. DPO directly learns from the preference data defined as $\mathcal{D} = \{(v_i, x_i, y^+_i, y^-_i)\}_{i=1}^N$, where the preference pair $(y^+_i, y^-_i)$ consists of a preferred and a dis-preferred response for the input $v_i$ and $x_i$. The optimization target is defined as:
% The preferences are quantified using a latent reward model $r(v, x, y)$, where the preferred response exhibits higher rewards and fewer hallucinations than the dis-preferred one.
% Following Bradley-Terry model \cite{bradley1952rank}, DPO formulates the probability of obtaining each preference pair as:
% \begin{align}
%     & p(y^+_i \succ y^-_i) = \sigma(r(v_i, x_i, y^+_i)-r(v_i, x_i, y^-_i)),
% \end{align}
% where $\sigma(\cdot)$ is the sigmoid function.
% DPO eliminates explicit reward modeling and derives the optimizing target as:
% \begin{align}
%     & \mathcal{L}_{DPO}(p_\theta;p_{ref})= - \mathbb{E}_{(v, x, y^+, y^-) \sim \mathcal{D}} \nonumber \\
%     & \Big[ \log \sigma \Big(\beta \log \frac{p_\theta(y^+|v, x)}{p_{ref}(y^+|v, x)}
%     - \beta \log \frac{p_\theta(y^-|v, x)}{p_{ref}(y^-|v, x)} \Big) \Big],
%     \label{formula:DPO}
% \end{align}
% \begin{figure*}[t]
% \centering
% \begin{equation}
%     \max_{\theta} \mathbb{E}_{(v_i, x_i, y^+_i, y^-_i) \sim \mathcal{D}}
%     \Big[ \log \sigma \Big(\beta (\log \frac{p_\theta(y^+_i|v_i, x_i)}{p_{ref}(y^+_i|v_i, x_i)}
%     - \log \frac{p_\theta(y^-_i|v_i, x_i)}{p_{ref}(y^-_i|v_i, x_i)} ) \Big) \Big],
%     \label{formula:DPO}
% \end{equation}
% \end{figure*}

% \begin{small}
{\small
\begin{align}
    & \max_{\theta} \mathbb{E}_{(v_i, x_i, y^+_i, y^-_i) \sim \mathcal{D}} \nonumber \\
    & \Big[ \log \sigma \Big(\beta \big( \log \frac{p_\theta(y^+_i|v_i, x_i)}{p_{ref}(y^+_i|v_i, x_i)}
    - \log \frac{p_\theta(y^-_i|v_i, x_i)}{p_{ref}(y^-_i|v_i, x_i)} \big) \Big) \Big],
    \label{formula:DPO}
\end{align}
}
% \end{small}
where $p_{ref}$ is defined by a reference model (usually fixed at the initial training checkpoint of $M_\theta$) and $\beta$ is a hyper-parameter to control the KL-divergence between $M_\theta$ and the reference model. 
% Intuitively, DPO increase the weighted relative log probability of preferred to dis-preferred responses, fostering outputs with reduced hallucinations.

\begin{figure*}[t]
	\begin{center}
		%\fbox{\rule{0pt}{2in} \rule{.9\linewidth}{0pt}}
		\includegraphics[width=0.95\linewidth]{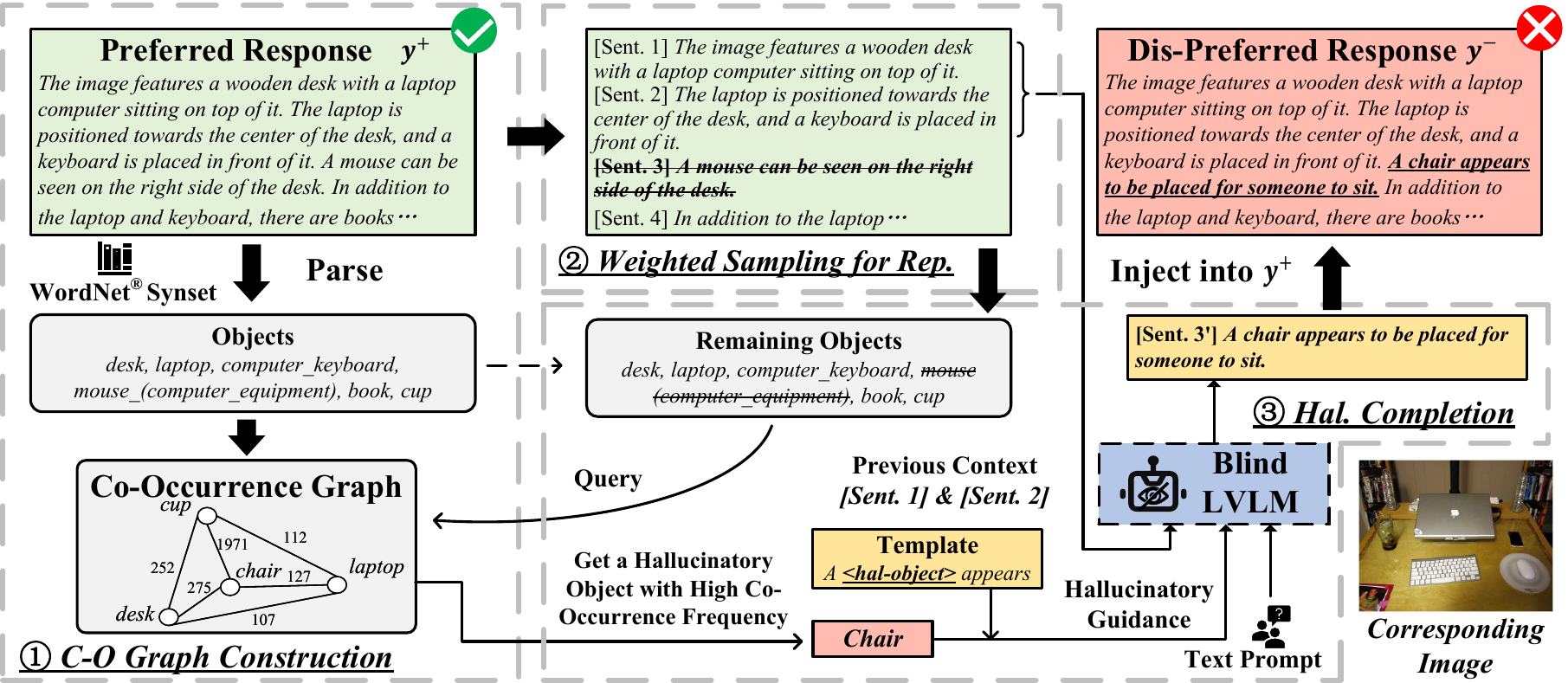}
	\end{center}
	\caption{
    Illustration of the self-injection pipeline: APASI first preprocesses the preferred responses to construct a graph documenting the co-occurrence relationships among objects. APASI then selects sentences to be replaced through weighted sampling, and employs the ``blind" target LVLM to make a hallucination completion for replacement with language-only input and hallucinatory guidance. The completed hallucinated sentence is injected into the preferred response to get the dis-preferred one.
    % Illustration of the self-injection pipeline.
    Best viewed in color.}
	\label{fig:data}
\end{figure*}

% \subsection{Preference Data Construction}
\subsection{Hallucination Self-Injection}
\label{ss:3.2}
The proposed APASI designs an autonomous preference data construction pipeline that uses only the target LVLM, without any need for additional annotation or auxiliary model. 
The data construction pipeline is predicated on the concept that a response $y^+_i$ which may already contain some hallucinations, and a response $y^-_i$ explicitly corrupted with additional hallucinations relative to $y^+_i$, form valid preference pair with varying preference levels.
Given an unannotated dataset $\mathcal{D}_{un}=\{(v_i,x_i)\}_{i=1}^{N}$ without ground-truth responses, APASI constructs preference pairs as:
\begin{itemize}
\item The \textbf{preferred} $y^+_i$: an original response generated by the target model $M_{\theta_0}$ with initial parameters $\theta_0$, i.e., $y^+_i \sim p_{\theta_0}(\cdot|v_i,x_i)$.
\item The \textbf{dis-preferred} $y^-_i$: a response that is created by deliberately injecting additional hallucinations into $y^+_i$.
\end{itemize}

% The self-injection process in APASI is based on three key observations of hallucinations:
% 1) LVLMs are prone to hallucinate objects that frequently \textbf{co-occur} with the existent objects in the image \cite{li2023evaluating};
% 2) LVLMs are prone to generate hallucinated contents with over-reliance on \textbf{language priors} \cite{favero2024multi};
% 3) More hallucinations tend to appear in the \textbf{latter parts} of the response \cite{zhou2024analyzing}. The target LVLM is guided to fabricate sentences about non-existent \textbf{co-occurring} objects with \textbf{language-only} inputs and then integrates this hallucinated content into the \textbf{latter parts} of the preferred response, forming the dis-preferred response.
% In this way, the preference pair facilitates an accurate learning signal for the mitigation of real hallucination patterns.

Specifically, to better simulate the real hallucination patterns in the dis-preferred response, APASI injects hallucination targeted to key observations of hallucinations including: 1) \textbf{Object co-occurrence}, 2) \textbf{Language prior}, and 3) \textbf{Positional factor}. 
We provide empirical evidence to support the importance of these observations in Appendix \ref{ss:c.halobsv}.
Accordingly, APASI guides the target model $M_{\theta_0}$ to generate sentences about non-existent \textbf{co-occurring} objects with \textbf{language-only} input and replaces the \textbf{latter part} of $y^+_i$ with these hallucinated sentences to obtain $y^-_i$. 
% Unlike existing preference data construction methods that depend on ranking feedback or correctional annotations from humans or external models, injecting additional hallucinations into responses is simpler and maintains the autonomy of the entire data construction pipeline. 
The detailed process is shown in Fig.\ref{fig:data} and is described below.

\subsubsection{Co-Occurrence Graph Construction}
% to simulate co-occurring hallucination -> find the occurence relationship underlying
% Before hallucination injection, APASI prepossesses generated preferred responses $\{y^+_i\}_{i=1}^{N}$ to determine the co-occurrence relationships among objects. 
To simulate co-occurring hallucination, it is necessary to uncover the biased co-occurrence relationships among objects inherent in the $M_{\theta_0}$ that exhibits such hallucinations.
APASI prepossesses the whole corpus of model-generated preferred responses $\{y^+_i\}_{i=1}^{N}$ to construct a co-occurrence graph $G$ documenting these relationships.
Each preferred response $y^+_i$ is first parsed into a set object tags $o_i$, using the WordNet \cite{miller1995wordnet} toolbox and synonym sets $S$. This process consolidates various synonyms into a single category tag for each object to simplify the analysis. APASI then uses these tags $\{o_i\}_{i=1}^{N}$ to build the co-occurrence graph $G$, where each node represents an object tag and the edge weight reflects the frequency of co-occurrence between the connected objects. Given an querying object, it's easy to get its co-occurring objects by traversing the nodes connected to the corresponding node.

\subsubsection{Weighted Sampling for Injection}
% APASI injects hallucinations into $y^+_i$ by replacing a percentage of $L$ original sentences, specified by the injection rate $\rho$, with hallucinated counterparts. 
% With an injection rate $\rho$, APASI injects hallucinations into $y^+_i$ by replacing a percentage of $\rho$ original sentences with hallucinated ones. 
% APASI injects hallucinations into $y^+_i$ by replacing a proportion $\rho$ (injection rate) of the $L$ original sentences with hallucinated counterparts. 
% Recognizing that hallucinations commonly occur at the latter part of the responses, APASI samples replacement indices from a multinomial distribution that assigns higher probabilities to later-positioned sentences in $y^+_i$. Details are in \textbf{Supplementary Material}.
APASI injects hallucinations into $y^+_i$ by replacing a proportion $\rho$ (injection rate) of the $L$ original sentences with hallucinated counterparts. The indices of the replaced sentences are sampled from a multinomial distribution defined by parameters $w_{1:L}$. Recognizing that hallucinations commonly occur in the latter part of the responses, we empirically set the last sentence twice as likely to be sampled as the first one. The weight of $k$-th sentence is defined as $w_k=1+\frac{k-1}{L}$ by linear interpolation.

% APASI injects hallucinations into $y^+_i$ by replacing the original sentences with hallucinated ones. 
% $y^+_i$ is decomposed into $L$ sentences like $y^+_i=(y^+_{i,1}, y^+_{i,2}, \dots, y^+_{i,L})$. With an injection rate of $\rho$, APASI uses weighted sampling to select $\rho L$ sentences for replacement, assigning higher weights to sentences positioned later in the text. 

\subsubsection{Hallucination Completion and Injection}
% APASI employs makes a hallucination completion about a non-exiting c for replacement with language-only input and hallucinatory guidance
Suppose APASI samples the $k$-th sentence in $y^+_{i}$ to be replaced with a hallucinated sentence $y^-_{i,k}$, resulting in the dis-preferred response:
\begin{align}
    & y^-_i=(y^+_{i,1}, \dots, y^+_{i,k-1}, y^-_{i,k}, y^+_{i,k+1}, \dots, y^+_{i,L}).
\end{align}
Considering the hallucination patterns of co-occurrence and language prior, the hallucinated $y^-_{i,k}$ is generated by a language-only ``blind'' LVLM under the guidance of a hallucinated co-occurring object $o_{i,k}^{hal}$, forming a description of $o_{i,k}^{hal}$. Specifically, $o_{i,k}^{hal}$ is obtained by querying the co-occurrence graph $G$ for an object that frequently co-occurs with objects in the remaining sentences, while ensuring $o_{i,k}^{hal}$ not in $o_i$. 
$o_{i,k}^{hal}$ is then put into a pre-defined guiding template, {\em e.g., ``A ($o_{i,k}^{hal}$) appears''}, which serves as hallucinatory guidance for the ``blind'' $M_{\theta_0}$ to make a hallucination completion $\tilde{y}_{i,k}$ as:
\begin{align}
    & \tilde{y}_{i,k} \sim p_{\theta_0} \big( \cdot|x_i, y^+_{i,<k}, temp(o_{i,k}^{hal}) \big),
\end{align}
where the text-only input includes the original prompt $x_i$, the previous context $y^+_{i,<k}$ of $k$-th sentence, and the filled template $temp(o_{i,k}^{hal})$. Note that if sentences after the $k$-th one are sampled to be replaced, the previous context includes $y^-_{i,k}$ instead of $y^+_{i,k}$. $y^-_{i,k}$ is finally obtained by composing the template and the completion as $y^-_{i,k}=(temp(o_{i,k}^{hal}), \tilde{y}_{i,k})$.

% The filled template $temp(o_{i,k}^{hal})$ serves as hallucinatory guidance for the model to generate $y^-_{i,k}$, which is composed of the template and a hallucination completion $\tilde{y}_{i,k}$, as $y^-_{i,k}=(temp(o_{i,k}^{hal}), \tilde{y}_{i,k})$.
% The visually disabled $M_{\theta_0}$ takes language-only inputs make a hallucination completion $\tilde{y}_{i,k}$ as:
% \begin{align}
%     & \tilde{y}_{i,k} \sim p_{\theta_0} \big( \cdot|x_i, y^+_{i,<k}, temp(o_{i,k}^{hal}) \big),
% \end{align}
% where the textual input includes the prompt $x_i$, the previous context $y^+_{i,<k}$ of $k$-th sentence, and $temp(o_{i,k}^{hal})$. 
% Note that if sentences after the $k$-th one are sampled to be replaced, the previous context includes $y^-_{i,k}$ other than $y^+_{i,k}$. 
% A complete description of the hallucination self-injection pipeline is provided in in \textbf{Supplementary Material}.

\subsection{Iterative Alignment with Curriculum Learning}
\label{ss:3.3}
% In the proposed preference collection pipeline, the model $M_{\theta_0}$ is static when generating preferred response generation and injecting hallucination. Simply running DPO on such offline collected preference data leads to distribution shift problem and causes sub-optimal alignment results \cite{gao2023scaling,yu2024rlaif}. However, collecting preference data at every optimization step brings complexity, resulting in difficulties in efficiently allocating computing resources. To strike a balance, APASI applies an iterative alignment strategy that alternates between data collection and DPO training epochs. 

% For training, APASI employs an iterative alignment strategy that cycles between data collection and model optimization to alleviate the distribution shift problem \cite{gao2023scaling,yu2024rlaif} and to promote continuous improvement.
% Moreover, to facilitate a more effective learning trajectory, the iterative strategy incorporates curriculum learning \cite{bengio2009curriculum}, which updates the preference data to be more challenging by increasing the hallucination injection rate over iterations. This tailored curriculum helps refine the model’s ability to distinguish the preference.
% using the latest optimized model

% brief intro for Iterative Alignment
To alleviate the distribution shift problem in preference alignment \cite{gao2023scaling,yu2024rlaif}, 
% APASI employs an iterative alignment strategy that alternates between model optimization and data collection using the latest optimized model. 
APASI employs an iterative alignment strategy. At iteration $t$, the latest optimized model $M_{\theta_{t-1}}$ is used for preference data construction including preferred responses generation and hallucination injection. $M_{\theta_{t-1}}$ is then optimized with DPO target in Equation (\ref{formula:DPO}) and preference data to get $M_{\theta_{t}}$ for data construction in the next iteration.

Furthermore, APASI incorporates a curriculum that progressively increases the difficulty of the alignment task, thereby facilitating a smoother and more effective learning trajectory over iterations \cite{bengio2009curriculum}. 
% The task's difficulty is inversely proportional to the gap within the preference pairs \cite{xu2023contrastive}. 
%  logic chain:
% 1. increased injection rate
% 2. narrowed gap
% 3. more challenging the task
% 4. increasing challenging of the task helps to refine the LVLM’s ability
The curriculum specifically reduces the injection rate $\rho$ with each iteration $t$, according to a monotonically decreasing curriculum function $f_c(t)$.
As $\rho$ decreases, the gap between the preferred response $y^+$ and the dis-preferred response $y^-$ narrows, intensifying the challenge of distinguishing subtle differences within the preference pairs \cite{xu2023contrastive}. 
This progressive increase in task difficulty is crucial for refining the LVLM's ability to accurately detect and reduce hallucinations.
The detailed algorithm for Iterative Alignment with Curriculum Learning outlined in Appendix \ref{ss:a.3}.

\begin{table*}[t]
\begin{center}
    \resizebox{1.82\columnwidth}{!}{
    \begin{tabular}{lcccccccc}
        \toprule[1.5pt]
        \multirow{3}{*}{\Large Model} & \multicolumn{5}{c}{Hallucination Benchmarks} & \multicolumn{3}{c}{Comprehensive Benchmarks} \\
        \cmidrule(r){2-6} \cmidrule(r){7-9}
        & \multicolumn{2}{c}{Object-Hal} & \multicolumn{2}{c}{AMBER} & POPE &
        \multirow{2}{*}{\begin{minipage}{2cm}\centering MMBench \\ TEST-v1.1 \end{minipage}} 
        & \multirow{2}{*}{MMVet} 
        & \multirow{2}{*}{\begin{minipage}{1.2cm}\centering LLaVA \\ BENCH \end{minipage}}\\
        \cmidrule(r){2-3} \cmidrule(r){4-5} \cmidrule(r){6-6}
        & CHAIR-s $\downarrow$ & CHAIR-i $\downarrow$ & CHAIR-i $\downarrow$  & F1 & F1 & & & \\ 
        \midrule
        LLaVA-v1.5-7B    & 51.0            & 13.7         & 7.8          & 74.7          & 85.9 (86.9)        & 62.3          & 30.5          & 63.4          \\ \midrule
        + POVID $^\dag$        & 33.6          & 9.0          & 5.2          & 86.5          & {\ul 86.9}  & 64.9          & 31.8          & 68.7          \\
        + HA-DPO $^\dag$       & -             & -              & 3.7          & 82.9          & 84.3        & -             & -             & 67.2          \\
        \midrule
        + RLAIF-V $^\ddag$      & \textbf{20.8} & {\ul 6.0}         & \textbf{2.8} & 84.5          & 78.9        & 63.6          & 30.1          & 64.9          \\
        + CLIP-DPO $^\ddag$      & -             & -              & 7.2          & 80.5          & 85.8        & -             & -             & -             \\
        % Self-Rewarding & -             & -              & -            & -             & {\ul 86.9}  & 64.5          & 31.4          & 61.2          \\
        + CSR $^\ddag$          & 28.0            & 7.5          & 4.4          & 86.5          & \textbf{87.0} & 65.4          & {\ul 33.9}    & {\ul 71.1}    \\
        \midrule
        + STIC $^\P$          & -             & -              & -            & -             & -           & 65.3          & 32.6          & 68.9          \\
        + SIMA $^\P$        & 41.6          & 13.0         & 6.6          & \textbf{86.9} & 85.8        & 64.9          & 31.6          & 66.1          \\ 
        \midrule
        + OPERA $^\S$        & 47.8          & 14.6           & -            & -             & 85.4        & 64.4          & -             & 60.3          \\
        + VCD $^\S$         & 48.6          & 14.9           & -            & 74.9          & 84.5        & -             & -             & 65.8          \\
        + Less is more $^\S$ & 36.8          & 11.3           & -            & 75.8          & 86.0          & -             & -             & 60.9          \\ \midrule
        + APASI-Base     & 38.1        & 9.2          & 6.0            & 86.1          & 85.6 (87.0)        & {\ul 66.7} & 33.5          & 67.3          \\
        + APASI-IACL     & 31.7        & 7.2          & 5.7          & 85.7          & 85.0 (87.0)          & 65.6    & \textbf{34.4} & \textbf{71.2} \\
        + APASI-Scaled   & {\ul 23.2}    & \textbf{5.1} & {\ul 3.5}    & {\ul 86.7}    & 85.0 (87.4)         & \textbf{67.2} & 32.2          & 70.1          \\     
        % \midrule  
        % LLaVA-v1.6-7B & 33.3 & 7.2 & 7.9 & 89.0   & 86.7 & 66.5 & 43.9 & 80.9 \\ 
        % STIC         & 32.0    & 9.1 & 4.7 & 87.2 & 82.9 & 67.8 & 45.0  & 79.2 \\
        % APASI-Base   &       &       &     &      &      &      &      &      \\
        % \midrule
        % Qwen2-VL-7B   & 44.4  & 9.473 & 6.7 & 89.9 & 87.5 & 80.7 & 62   & 92.3    \\
        % APASI-Base   & \textbf{23.8} & \textbf{6.2} & \textbf{5.8} & 89.4 & \textbf{87.6} & 79.4 & \textbf{63.7} & \textbf{93.5}    \\
        \bottomrule[1.5pt]
    \end{tabular}
    }
    % }
\end{center}
\caption{Performance of APASI with the LLaVA-v1.5-7B baseline compared with other hallucination mitigation methods across various benchmarks. 
$\dag$, $\ddag$, $\P$, and $\S$ respectively indicate RLAIF methods with proprietary models, RLAIF methods with open-source models, self-improvement methods, and non-alignment-based methods.
}
\label{tab:main}
\end{table*}

\section{Experiments}
\subsection{Experimental Setups}
% In this subsection, we provide the experimental settings of APASI, with more details available in the \textbf{Supplementary Material}.
\paragraph{Preference Data.} 
The construction of preference data in APASI leverages self-injection based on the detailed description task, which requires LVLMs to accurately perceive and describe visual elements, thus directly reflecting hallucination issues. We construct the \textbf{SI-23k} dataset derived from images and descriptive responses in the {\em detail-23k} subset of the LLaVA's instruction tuning dataset, excluding ground-truth responses. We further construct the scaled-up \textbf{SI-130k} by adding unannotated images from the VisualGenome (VG) dataset \cite{krishna2017visual}. The descriptive prompts in SI-130k are from SI-23k. For object parsing, we employ LVIS object synonym sets \cite{gupta2019lvis}, which are based on WordNet. The default injection rate $\rho$ is set to 0.2.

\paragraph{Implementation Details.} 
In this study, we choose the supervised fine-tuned LLaVA-v1.5-7B \cite{liu2024improved}, a widely used baseline, as the target model for hallucination mitigation. 
% To further verify APASI’s compatibility across different LVLMs, we also include LLaVA-vicuna-v1.6-7B \cite{liu2024llava} and Qwen2-VL-7B \cite{wang2024qwen2} as additional baseline models.
Unless specified otherwise, all experiments are conducted with LLaVA-v1.5-7B and SI-23k. We use SI-23k and SI-130k to train the target LVLM in a single iteration to obtain \textbf{APASI-Base} and \textbf{APASI-Scaled}, respectively. For iterative alignment with curriculum learning (IACL), we train the LVLM for $T=3$ iterations with a curriculum function $f_c(t)=0.8-0.2t$ and get \textbf{APASI-IACL}. A single iteration takes about 520 minutes running with 8 V100 GPUs and data construction accounts for about 31.2$\%$ of time. 

\paragraph{Evaluation.}
% benchmarks
For quantitative analysis, we conduct evaluation on both hallucination benchmarks (Object-Hal \cite{rohrbach2018object}, AMBER \cite{wang2023llm}, and POPE \cite{li2023evaluating}) and comprehensive benchmarks (MMBench \cite{liu2025mmbench}, MMVet \cite{yu2023mm}, and LLaVABench \cite{liu2024visual}). Object-Hal provides CHAIR-i (C-i) and CHAIR-s (C-s) metrics measuring the ratio of hallucinated objects and responses respectively in generative tasks. POPE uses F1 scores in discriminative tasks. AMBER reports both C-i and F1. We report overall scores for the comprehensive benchmarks. 
% We also select samples in the test2017 split of MSCOCO \cite{lin2014microsoft} for analysis.

\subsection{Main Results}
% improvement on baseline
% hal and comprehensive

% detail 23k: which is used in SFT: necessity of preference alignment
% iacl and data scale up

% comparing with sota
% surpass the self-improvement and not based on alignment
\subsubsection{Performance Comparison with the Baseline Model}
Results show that APASI effectively mitigates the hallucination problem in the LLaVA-v1.5-7B baseline model, as shown in Tab.\ref{tab:main}.
Specifically, APASI-Base reduces the ratio of hallucinated objects by 4.5/1.8 in generative tasks on Object-Hal and AMBER, respectively, and reduces the ratio of hallucinated responses by 12.9 on Object-Hal. APASI-Base also improves the performance on all three comprehensive benchmarks for the baseline. 
Further incorporating IACL makes improvement on most of the benchmarks, suggesting the effectiveness of IACL.
Notably, both APASI-Base and APASI-IACL are trained with SI-23k, where the images and the textual prompts are already used in the supervised fine-tuning of the baseline. The enhancement brought about by seen data indicates the effectiveness of the preference-alignment training paradigm in deepening the LVLM's utilization of existing data.
Scaling up preference data also takes effect, especially in reducing hallucinations. APASI-Scaled with SI-130k further reduces hallucination ratios of the baseline by 27.8/8.6/4.3 on Object-Hal and AMBER.

In contrast to generative performance, APASI shows inconsistent discriminative results, particularly under-performing the baseline on POPE.
% This can be explained by the fact that the alignment goal in APASI is set to reduce hallucination in generations, while the discriminant ability has not been explicitly optimized. 
This discrepancy stems from APASI's alignment goal, which does not directly optimize discriminative capabilities.
To better ground discriminative ability to the hallucination-reduced descriptions, we prompt the LVLMs with {\em Describe the image and answer the question.} during POPE testing. The modified F1 scores, displayed next to the original scores in Table \ref{tab:main}, show improvements: the baseline increases by $1.16\%$, while APASI-Base/IACL/Scaled improve by $1.64\%/2.35\%/2.82\%$, These results indicates that the LVLMs with better generative performance exhibit greater improvement.

\subsubsection{Performance Comparison with the SOTA Methods}
% sota
We also compare APASI with the SOTA hallucination mitigation methods in Tab.\ref{tab:main}. The methods are categorized as follows: 1) RLAIF methods with proprietary models: POVID \cite{zhou2024aligning} and HA-DPO \cite{zhao2023beyond}; 2) RLAIF methods with open-source models: RLAIF-V \cite{yu2024rlaif}, CLIP-DPO \cite{ouali2025clip}, and CSR \cite{zhou2024calibrated}; 3) self-improvement methods: STIC \cite{deng2024enhancing} and SIMA \cite{wang2024enhancing}; 4) methods without preference alignment: OPERA \cite{huang2024opera}, VCD \cite{leng2024mitigating} and Less is more \cite{yue2024less}. Methods in categories 1) and 2) rely on external resources, whereas those in 3) and 4) do not.

Without the need for any external support, APASI achieves comparable or even better performance across all three hallucination benchmarks compared to methods dependent on external resources. APASI also outperforms all methods without external dependencies on four out of five hallucination metrics. 
For comprehensive abilities, APASI performs the best across all three benchmarks, highlighting its effectiveness in both hallucination mitigation and overall ability enhancement.
% To summarize, APASI achieves considerable performance on both hallucination mitigation and comprehensive ability enhancement without external dependency, showing the superiority of our approach.

\subsection{Ablation Studies and Analysis}

% Qwen2-vl-7B is recognized as one of the most advanced open-source models within the 7B parameter range

% \begin{table}[h]
% \begin{center}
%     \begin{tabular}{lccc}
%         \toprule[1.5pt]
%         \multirow{2}{*}{\Large Prompt} 
%         & Default & \multicolumn{2}{c}{Descriptive}  \\
%         \cmidrule(r){2-2} \cmidrule(r){3-4}
%         & F1 & F1 & $\Delta$ \\ 
%         \midrule
%         LLaVA-v1.5-7B & 85.9 & 86.9          & +1.16\%          \\
%         \midrule
%         APASI-Base    & 85.6 & 87.0            & +1.64\%          \\
%         APASI-IACL    & 85.0   & 87.0            & +2.35\%          \\
%         APASI-Scaled  & 85.0  & \textbf{87.4} & \textbf{+2.82\%} \\
%         \bottomrule[1.5pt]
%     \end{tabular}
% \end{center}
% \caption{pope}
% \label{tab:pope}
% \end{table}
\subsubsection{Ablation Studies on Injection Settings} 
As shown in Tab.\ref{tab:setting}, we design ablative experiments to verify the effectiveness of the settings in getting the dis-preferred responses trough self-injection. We specifically compare APASI-Base with five variants when injecting hallucinations. Qualitative analyses are in Appendix \ref{ss:c.1}.

Results show that it will bring performance drop if using random non-existent objects instead of co-occurring ones for hallucinatory guidance. Further removing the guiding template for completion even causes the model to underperform the baseline. This is because the language-only model may not reliably generate hallucinated sentences only based on textual prompt and the previous context, making the preference pairs invalid. These results verify the effectiveness of the hallucinatory guidance with co-occurring objects.

It will also bring performance drop if injecting hallucinations by replacing with the co-occurring object word, instead of a sentence describing the object. 
Only replacing words may result in the injected target words being markedly inconsistent with the context. Such dis-preferred responses with ridiculous mistakes fail to provide effective learning signal, verifying the importance of employing the language-only model for injection to maintain reasonableness in the dis-preferred responses.

The decline observed from removing weighted sampling when deciding sentences to be replaced shows the necessity of considering positional factor in self-injection. Using the ground-truth responses instead of the model-generated ones as the preferred responses to be injected does not result in performance changes. However, this will limit APASI to scale up to unannotated data.
 
% 1) using random non-existent objects instead of co-occurring ones for hallucinatory guidance;
% 2) removing hallucinatory guidance when generating injected sentences; 
% 3) only replacing with co-occurring object word instead of the whole fabricated sentence.
% 4) removing weighted sampling when deciding sentences to be replaced;
% We also implement a variant using the ground-truth responses as the preferred responses to be injected. 

\begin{table}[t]
\begin{center}
    \resizebox{0.7\columnwidth}{!}{
    \setlength{\tabcolsep}{1.2mm}{
    \begin{tabular}{lccc}
        \toprule[1.5pt]
        \multirow{2}{*}{\Large Setting} 
        % & \multirow{2}{*}{\Large $\rho$} 
        & \multicolumn{2}{c}{Object-Hal} & AMBER \\
        \cmidrule(r){2-3} \cmidrule(r){4-4}
        &  C-s $\downarrow$ & C-i $\downarrow$ & C-i $\downarrow$ \\ 
        \midrule
        LLaVA-v1.5-7B    & 51.0            & 13.7         & 7.8 \\
        APASI-Base            & 38.1 & 9.2  & 6.0  \\
        \midrule
        random guide          & 44.4 & 12.2 & 7.0   \\
        w/o guide             & 66.7 & 21.0  & 11.3 \\
        replace object        & 49.6   & 13.5 & 8.1  \\
        w/o w.s.              & 39.7 & 9.7  & 6.0  \\
        \midrule
        GT as preferred        & 38.1 & 9.7  & 5.9  \\
        \bottomrule[1.5pt]
    \end{tabular}
    }
    }
\end{center}
\caption{Ablation studies on injection settings.}
\label{tab:setting}
\end{table}

% \begin{table}[t]
% \begin{center}
%     \begin{tabular}{cccc}
%         \toprule[1.5pt]
%         \multirow{2}{*}{\Large $\rho$} 
%         & \multicolumn{2}{c}{Object-Hal} & AMBER \\
%         \cmidrule(r){2-3} \cmidrule(r){4-4}
%         & C-i $\downarrow$ & C-s $\downarrow$ & C-i $\downarrow$ \\ 
%         \midrule
%         0.1         & 49.2           & 15.3         & 8.7 \\
%         \textbf{0.2} & \textbf{38.1} & \textbf{9.2} & \textbf{6.0} \\
%         0.3          & \textbf{38.1} & 10.5         & 6.1        \\
%         0.4          & 42.4            & 11.2         & 6.2        \\
%         0.5          & 41.3           & 10.9         & 6.2        \\
%         0.6          & 46.4            & 12.3         & 6.4       \\
%         \bottomrule[1.5pt]
%     \end{tabular}
% \end{center}
% \caption{Ablation studies on injection rates.}
% \label{tab:rate}
% \end{table}

\subsubsection{Analysis on IACL and Sustainability of Improvement.}
\begin{figure}[t]
	\begin{center}
		%\fbox{\rule{0pt}{2in} \rule{.9\linewidth}{0pt}}
		\includegraphics[width=\linewidth]{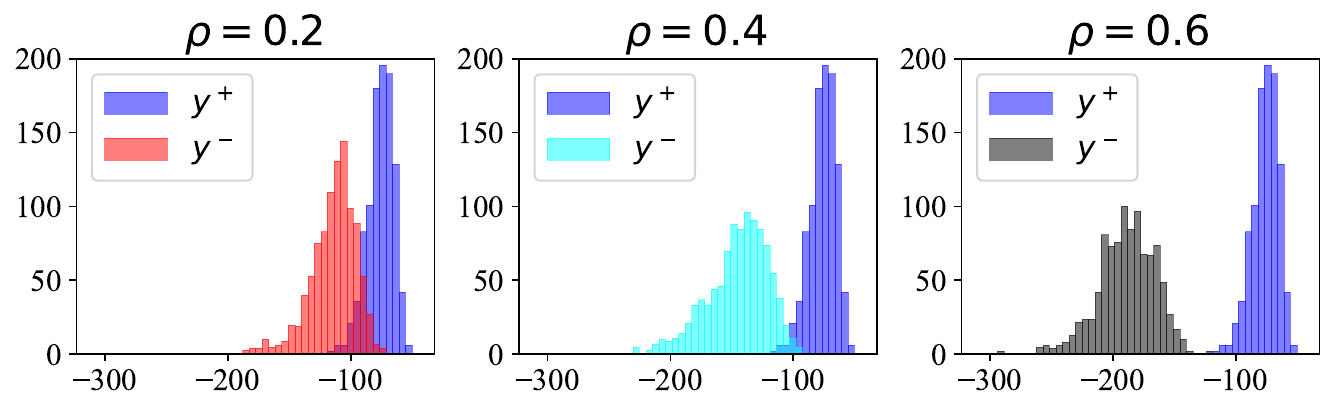}
	\end{center}
	\caption{Comparative histograms of preferred vs dis-preferred log probabilities with different injection rates. Best viewed in color.}
	\label{fig:logp}
\end{figure}

We first select 1000 samples from SI-23k to plot the histograms of the preferred and dis-preferred log probabilities with different injection rates in Fig.\ref{fig:logp}. Results show that as the injection rate decreases, the dis-preferred responses are more likely to be generated and less easily distinguishable by the target LVLM, thus making the alignment task more challenging. This observation supports our strategy of employing a curriculum with decreasing injection rate.

\begin{figure}[t]
	\begin{center}
		%\fbox{\rule{0pt}{2in} \rule{.9\linewidth}{0pt}}
		\includegraphics[width=\linewidth]{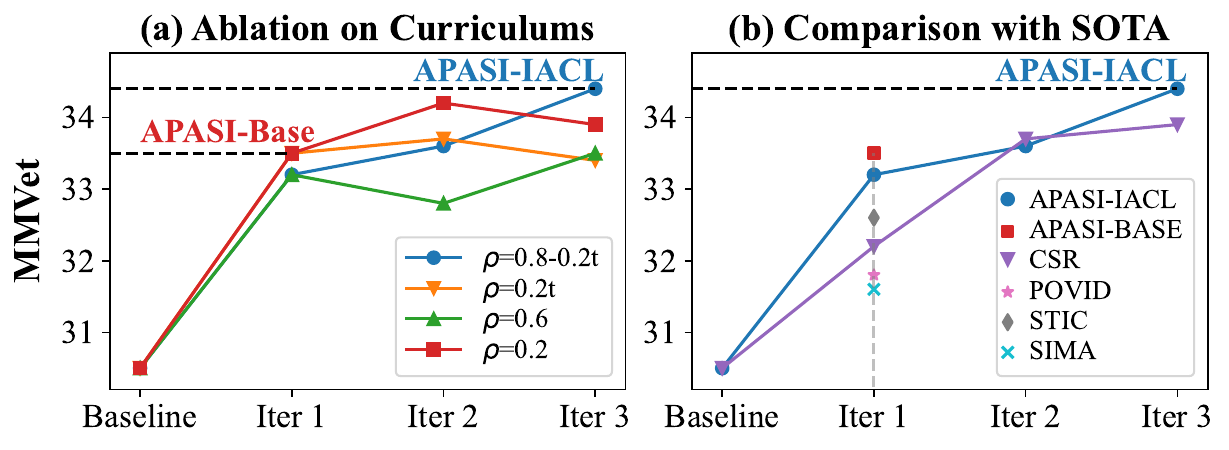}
	\end{center}
	\caption{MMVet scores of: \textbf{(a)} APASI at each iterations with different curriculums for injection rate; \textbf{(b)} different models at each iterations.
    Best viewed in color.}
	\label{fig:ia}
\end{figure}

To evaluate the effectiveness of iterative alignment with curriculum learning, we plot performance on MMVet across iterations in Fig.\ref{fig:ia}(a). Besides APASI-IACL with a decreasing injection rate $\rho$ (0.6 to 0.2), we evaluate three variants: 1) a reverse curriculum where $\rho$ increases from 0.2 to 0.6; 2) $\rho$ fixed at 0.6; 3) $\rho$ fixed at 0.2. 
% In a single alignment round, methods with a lower $\rho$ outperform those with a higher $\rho$, indicating that more challenging preference pairs enhance performance within a single iteration.
In single-round alignment, lower $\rho$ yields better performance, suggesting harder preference pairs promote learning.
% This verifies that a closer gap between the preference pairs makes the alignment task more challenging, thus brought more performance gain in a single iteration. 
Detailed results for different $\rho$ are in the Appendix \ref{ss:b.2}.
Over iterations, APASI with fixed $\rho$ and the reverse curriculum show fluctuating improvements, while APASI-IACL with decreasing $\rho$ demonstrates a smooth trajectory and achieves the best performance at iter3 with a $3.6\%$ gain on MMVet compared to iter1.
These results verify that a well-designed curriculum fosters smooth learning to benefit sustainable improvement over iterations.

We further compare APASI-IACL with SOTA methods in \ref{fig:ia}(b). Most methods only perform a single-round alignment and are outperformed by APASI-IACL. 
CSR \cite{zhou2024calibrated} supports iterative alignment. However, CSR's reliance on an external model for preference data collection limits its effectiveness, thus it yields inferior results to the autonomous APASI-IACL. These results validate APASI's advantage in facilitating sustainable model improvement. 

\begin{table}[t]
\begin{center}
    \resizebox{0.8\columnwidth}{!}{
    \setlength{\tabcolsep}{0.8mm}{
    \begin{tabular}{lcccc}
        \toprule[1.5pt]
        \multirow{2}{*}{\Large Model} 
        & \multicolumn{2}{c}{Object-Hal} 
        & \multirow{2}{*}{MMVet} 
        & \multirow{2}{*}{\begin{minipage}{1.2cm}\centering LLaVA \\ BENCH \end{minipage}}\\
        \cmidrule(r){2-3}
        & C-s $\downarrow$ & C-i $\downarrow$ & & \\ 
        \midrule
         LLaVA-v1.6-7B & 38.1 & 8.8 & 42.5 & 75.6 \\ 
        % + STIC         & 32.0    & 9.1 & 45.0  & 79.2 \\
        + APASI-Base   & \textbf{28.8} & \textbf{7.2} & \textbf{44.2} & \textbf{80.4}      \\
        \midrule
        Qwen2-VL-7B   & 44.4  & 9.5 & 62.0   & 92.3    \\
        + APASI-Base   & \textbf{23.8} & \textbf{6.2} & \textbf{63.7} & \textbf{93.5}    \\
        \bottomrule[1.5pt]
    \end{tabular}
    }
    }
\end{center}
\caption{Performance of APASI with LLaVA-v1.6-7B and Qwen2-VL-7B baselines.}
\label{tab:model}
\end{table}

\begin{figure*}[t]
	\begin{center}
		%\fbox{\rule{0pt}{2in} \rule{.9\linewidth}{0pt}}
		\includegraphics[width=0.95\linewidth]{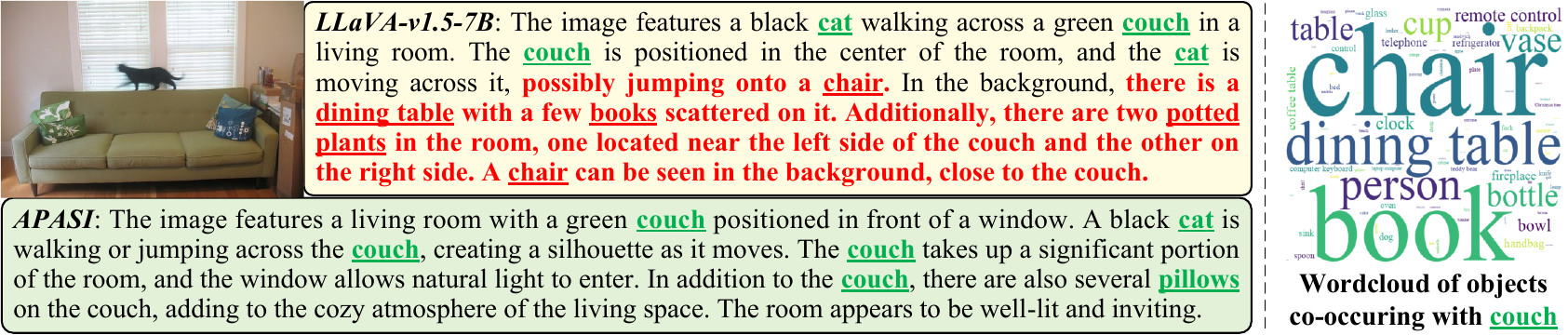}
	\end{center}
	\caption{\textbf{Left}: Comparison of baseline and APASI responses. Correct and incorrect objects are underlined in red and green respectively. Other hallucinated contents are in red. \textbf{Right}: The wordcloud of objects co-occurring with {\ul couch} obtained from baseline's generations.}
	\label{fig:case}
\end{figure*}

\begin{figure}[t]
	\begin{center}
		%\fbox{\rule{0pt}{2in} \rule{.9\linewidth}{0pt}}
		\includegraphics[width=\linewidth]{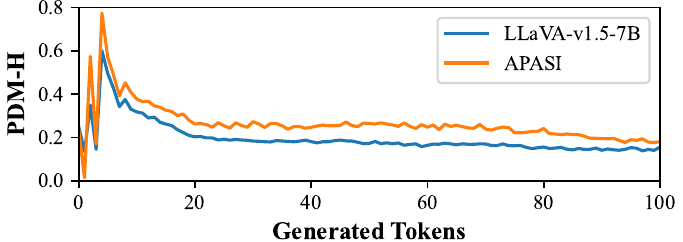}
	\end{center}
	\caption{Averaged PDM-H for responses generated by the baseline and APASI. Best viewed in color.}
	\label{fig:pdmh}
\end{figure}

\subsubsection{Analysis on generalization capability}
To assess APASI's applicability across different LVLMs, we apply APASI to two additional baseline models: LLaVA-v1.6-7B(vicuna) \cite{liu2024llava}, an improved version of LLaVA-v1.5-7B, and Qwen2-VL-7B \cite{wang2024qwen2}, which has an different architecture from the LLaVA series and is among the most advanced open-source LVLMs. 
Note that LLaVA-v1.6-7B and Qwen2-VL-7B autonomously generate their versions of the SI-23k dataset for preference alignment. 
Results in Tab.\ref{tab:model} show that even these advanced LVLMs are not entirely unaffected by hallucinations. APASI effectively mitigates this problem and improves overall performance on comprehensive benchmarks for both LLaVA-v1.6-7B and Qwen2-VL-7B.
These results demonstrate the APASI's robust generalization capabilities, particularly its compatibility with difference model architectures.

\subsubsection{Analysis on Reliance of Language Priors}
To assess the reliance on \textbf{language priors}, a key hallucination pattern, we calculate the Prompt Dependency Measure based on Hellinger distance (PDM-H) \cite{favero2024multi}. 
PDM-H quantifies the divergence in the LVLM's probability distribution when generating a token under image-language input versus language-only input.  
% It is calculated as at $j$-th step:
% \begin{align}
%     & \text{PDM-H}(j)=H \big(p(\cdot|v,x,y_{<j}),p(\cdot|x,y_{<j})\big),
% \end{align}
% The Hellinger distance is defined as:
% \begin{align}
%     & H(p, q) = \frac{1}{\sqrt{2}} \sqrt{\sum_{i=1}^d (\sqrt{p_i} - \sqrt{q_i})^2},
% \end{align}
% where $p = (p_1, p_2, \ldots, p_d)$ and $Q = (q_1, q_2, \ldots, q_d)$ are two $d$-dimension discrete probability distributions.
% PDM-H measures the difference of LVLM's generative probability distribution under image-language input and language-only input. 
A lower PDM-H indicates greater reliance on the textual input. We randomly sample 1000 images from COCO-test2017 \cite{lin2014microsoft} and generate detailed descriptions using the LLaVA-v1.5-7B baseline and APASI respectively. The averaged PDM-H curves, depicted in Fig.\ref{fig:pdmh}, show that APASI exhibits higher PDM-H than the baseline. This result indicates APASI's effectiveness in mitigating the over-reliance on \textbf{language priors}. More details about PDM-H are given in \ref{ss:a.5}.

\subsection{Qualitative Analysis}
To intuitively show the effectiveness of APASI in mitigating hallucinations, we compare responses from LLaVA-v1.5-7B baseline and APASI in Fig.\ref{fig:case}. 
Both models are given with the same image from COCO-test2017 with a detailed description prompt. The response of the baseline exhibited significant hallucinations with non-existent objects in the latter part. Notably, incorrect objects like {\ul dining table}, {\ul chair}, and {\ul book}. usually appears in indoor scenes together with {\ul couch}. This pattern of co-occurrence is further verified by the wordcloud of \textbf{co-occurring} objects in Fig.\ref{fig:case}, which derived from statistical analysis of the baseline's outputs. Remarkably, APASI effectively eliminates the \textbf{co-occurring} hallucination, while capturing all major objects in the image. 
% More examples are presented in \textbf{Supplementary Material}.
% To analyze APASI's handling of {\em over-reliance on language priors}, another key hallucination pattern, we calculate Prompt Dependency Measure based on the Hellinger distance (PDM-H), following \cite{favero2024multi}. 

\section{Conclusions}

% In this paper, we have proposed APASI, a novel method designed to mitigate hallucinations in LVLMs through preference alignment. 
% APASI has got rid of limitations from external dependency through self-injecting hallucinations into model-generated responses to construct preference data.
% APASI has overcome limitations from external dependencies by autonomously injecting hallucinations into model-generated responses to construct preference data.
We propose APASI, a novel method for mitigating hallucinations in LVLMs via preference alignment without external dependencies. 
This self-injecting process, based on key observations, has provided an accurate learning signal for effective preference alignment.
APASI has further employed iterative alignment with curriculum learning for improving the training process.
% Extensive experiments have demonstrated our effectiveness and superiority in hallucination mitigation across three baselines and various benchmarks.
Extensive experiments have demonstrated our effectiveness and superiority across three baselines and various benchmarks.
% APASI has successfully mitigated hallucination problem for three baseline models and achieved comparable or even superior performance to methods with external dependency. Extensive quantitative and qualitative experiments on six benchmarks have demonstrated the effectiveness of APASI. 

\section*{Limitations}
% Due to computational resource constraints, APASI has only been applied to models with limited amount of parameter (e.g. 7B models). However, in real-world applications, larger-scale models are becoming increasingly prevalent. We will explore applying APASI to larger models to evaluate its generalization capability and scalability in future work. 
Our experiments were conducted under computational resource constraints, which restricted the application of APASI to relatively small-scale models (e.g., 7B parameters). In contrast, real-world deployments increasingly rely on larger models, and assessing APASI’s effectiveness in such settings remains an important direction for future work.
Additionally, similar to many prior studies, APASI primarily addresses inconsistencies between visual inputs and generated textual responses, i.e., fidelity-related hallucinations. While our results suggest that APASI also yields improvements on knowledge-intensive tasks, hallucinations stemming from factual inaccuracies about real-world knowledge are not explicitly modeled in the current framework. Addressing this limitation will be the focus of future research.
 
\section*{Acknowledgments}
This work is supported by Beijing Natural Science Foundation (L243015, L223003), the Natural Science Foundation of China (No. 62036011, 62192782), the Project of Beijing Science and Technology Committee (No. Z231100005923046).

\bibliography{acl_latex}

\appendix

\section{Details about the Method and Implementation}

\subsection{Details of Hallucination Self-Injection}
For a preferred response with $L$ sentences and an injection rate of $\rho$, the number of the sentence to be replaced is rounded $\rho L$. If the number is zero, the sample will be discarded.
Note that the first sentence in the preferred response is excluded from weighted sampling, to avoid insufficient previous text for hallucination completion.

The average sentence number is $5.5$ for responses generated by LLaVA-v1.5-7B on the source dataset of LLaVA-detail-23k \cite{liu2024visual}. The corresponding SI-23k dataset with $\rho$ set to 0.2 contains 23,196 samples after discarding invalid samples.

\begin{table}[t]\centering
\begin{minipage}{0.95\columnwidth}\vspace{0mm}    \centering
\begin{tcolorbox} 
    \centering
    \small
     \hspace{-6mm}
\begin{itemize}
\item ``A \textless hal-object\textgreater appears''
\item ``There is a \textless hal-object\textgreater''
\item ``There are \textless hal-object\textgreater''
\item ``\textless hal-object\textgreater can also be seen''
\item ``\textless hal-object\textgreater can be seen''
\item ``You can see a \textless hal-object\textgreater''
\item ``There are multiple \textless hal-object\textgreater''
\item ``Several \textless hal-object\textgreater can be observed''
\item ``Some \textless hal-object\textgreater are present''
\item ``Among the items, there is a \textless hal-object\textgreater''
\item ``In the image, there is a \textless hal-object\textgreater''
\item ``On the right, there is a \textless hal-object\textgreater''
\item ``On the left, a \textless hal-object\textgreater is present''
\item ``In the center, you see a \textless hal-object\textgreater''
\item ``At the top, there is a \textless hal-object\textgreater''
\item ``At the bottom, a \textless hal-object\textgreater is visible''
\item ``In the background, a \textless hal-object\textgreater can be seen''
\item ``In the foreground, there is a \textless hal-object\textgreater''
\item ``To the side, a \textless hal-object\textgreater is located''
\item ``Near the edge, a \textless hal-object\textgreater appears''
\item ``Close to the center, a \textless hal-object\textgreater is seen''
\end{itemize}
\end{tcolorbox}
\caption{The list of hallucinatory guiding templates.}
    \label{tab:templates}
\end{minipage}
\end{table}

\begin{algorithm*}[t]
    \caption{APASI incorporating iterative alignment with curriculum learning}
    \textbf{Input}: Unannotated dataset: $\mathcal{D}_{un}=\{(v_i,x_i)\}_{i=1}^{N}$. Initial target LVLM: $M_{\theta_0}$. Synonym sets for objects: $S$. Curriculum function for injection rate: $f_{c}(\cdot)$. Number of iterations: $T$. \\
    \textbf{Output}: $M_{\theta_T}$  with reduced hallucination 
    
    \begin{algorithmic}[1] %[1] enables line numbers
        \For{$t=1,\dots,T$}
        \State $\{y^+_i\}_{i=1}^{N} \leftarrow \text{Generation}(M_{\theta_{t-1}}, \mathcal{D}_{un})$
        \State $\{o_i\}_{i=1}^{N}, G \leftarrow \text{Preprocessing}(\{y^+_i\}_{i=1}^{N}, S)$
        \State $\rho=f_{c}(t)$
        \State $\{y^-_i\}_{i=1}^{N} \leftarrow \text{HalInjection}
        (M_{\theta_{t-1}}, \{y^+_i\}_{i=1}^{N}, \{o_i\}_{i=1}^{N}, G, \rho)$
        % \Statex $\phantom{\{y^-_i\}_{i=1}^{N} \leftarrow \text{HalInjection}aaa} \{o_i\}_{i=1}^{N}, G, \rho)$
        \State $\mathcal{D} \leftarrow \{(v_i, x_i, y^+_i, y^-_i) \mid i = 1 \text{ to } N\}$
        \State $M_{\theta_t} \leftarrow \text{DPOTraining}(M_{\theta_{t-1}}, \mathcal{D})$
        \EndFor
        \State \textbf{return} $M_{\theta_T}$    
    \end{algorithmic}
    \label{algo:1}
\end{algorithm*}

\subsection{Hallucinatory Guiding Templates}

APASI employs guiding templates to guide the target LVLM to fabricate content about non-existent co-occurring objects for hallucination self-injection. We pre-define the templates according to the usual structure of descriptive sentence, as listed in Tab.\ref{tab:templates}. During every self-injection, one of the templates is randomly sampled and instantiated by filling in the co-occurring object \textless hal-object\textgreater. The target LVLM then follows the template to make a hallucinatory completion.

\subsection{Details of Iterative Alignment with Curriculum Learning}
\label{ss:a.3}
As outlined in Algorithm \ref{algo:1}, at each iteration, the preferred responses $\{y^+_i\}_{i=1}^{N}$ are generated using the latest model $M_{\theta_{t-1}}$ and are preprocessed to get the object tags $\{o_i\}_{i=1}^{N}$ and the co-occurrence graph $G$. 
Following this, $M_{\theta_{t-1}}$ inject hallucinations into $\{y^+_i\}_{i=1}^{N}$ to get the dis-preferred responses $\{y^-_i\}_{i=1}^{N}$, with a injection rate $\rho=f_{c}(t)$ determined by the curriculum. 
% The preference dataset $\mathcal{D}$ is constructed.
Subsequently, $M_{\theta_{t-1}}$ is trained with DPO optimization target with the preference dataset $\mathcal{D}$, resulting in the optimized $M_{\theta_t}$ for data construction in the next iteration. 
In this way, APASI operates smoothly in an iterative manner, ensuring sustainable improvement.
% , resulting in better learning efficiency.

\subsection{Descriptive Prompts}

In the experiments, we scale-up the preference data to images from the VisualGenome (VG) \cite{krishna2017visual}, where the textual prompts are provided for the images. For each image in VG, we simply pair it with a random descriptive prompt in SI-23k, which is sourced from the detail-23k subset of the LLaVA’s instruction tuning dataset \cite{liu2024visual}. The prompts are listed in Tab.\ref{tab:descriptive_prompts}

\begin{table}[h!]\centering
\begin{minipage}{0.95\columnwidth}\vspace{0mm}    \centering
\begin{tcolorbox} 
    \centering
    \small
     \hspace{-6mm}
\begin{itemize}
\item ``What is this photo about?"
\item ``Explain the visual content of the image in great detail."
\item ``Describe the following image."
\item ``What do you see happening in this image?"
\item ``Analyze the image in a comprehensive and detailed manner."
\item ``Write a detailed description of the given image."
\item ``What's happening in the scene?"
\item ``What do you think is going on in this snapshot?"
\item ``Can you elaborate on the elements of the picture provided?"
\item ``What are the key elements in this picture?"
\item ``Can you describe the main features of this image for me?"
\end{itemize}
\end{tcolorbox}
\caption{The list of descriptive prompts used in preference data construction.}
    \label{tab:descriptive_prompts}
\end{minipage}
\end{table}

\subsection{Prompt Dependency Measure}
\label{ss:a.5}
We calculate the Prompt Dependency Measure based on Hellinger distance (PDM-H) \cite{favero2024multi} measuring the difference of LVLM's generative probability distribution when given image-language input and language-only input. 
It is calculated at the $j$-th step as follows:
\begin{align}
    & \text{PDM-H}(j)=H \big(p(\cdot|v,x,y_{<j}),p(\cdot|x,y_{<j})\big),
\end{align}
The Hellinger distance is defined as:
\begin{align}
    & H(p, q) = \frac{1}{\sqrt{2}} \sqrt{\sum_{i=1}^d (\sqrt{p_i} - \sqrt{q_i})^2},
\end{align}
where $p = (p_1, p_2, \ldots, p_d)$ and $Q = (q_1, q_2, \ldots, q_d)$ are two $d$-dimension discrete probability distributions.

\subsection{Implementation Details of Training}
During the alignment training, we employ LoRA \cite{hu2021lora} for efficient tuning. Visual encoders of all three models are frozen.
The hyperparameters for different baseline models are listed in Tab.\ref{tab:appendix_setting}. All experiments are conducted on 8 V100 32GB GPUs. It takes about 6/8/8 hours for training LLaVA-v1.5-7B, LLaVA-v1.6-7B, and Qwen2-VL-7B respectively.

\begin{table}[t]
\resizebox{1\columnwidth}{!}{
\setlength{\tabcolsep}{0.8mm}{
\centering
\begin{tabular}{cccc}
\toprule[1.5pt]
\multirow{2}{*}{\begin{minipage}{2cm}\centering Baseline \\ Model \end{minipage}} &
\multirow{2}{*}{LLaVA-v1.5-7B} &
\multirow{2}{*}{LLaVA-v1.6-7B} &
\multirow{2}{*}{Qwen2-VL-7B} \\
& & & \\
\midrule
$\beta$ & 0.1 & 0.1 & 0.1 \\
\#epoch        & 3             & 1             & 1           \\
learning rate  & 4e-7          & 1e-6          & 1e-6        \\
batchsize      & 64            & 64            & 64          \\
lora\_r        & 128           & 64            & 64          \\
lora\_alpha    & 256           & 128           & 128         \\
\bottomrule[1.5pt]
\end{tabular}
\centering
}}
\caption{Training configurations for different baseline models.}
	\label{tab:appendix_setting}
\end{table}

\begin{table}[t]
\resizebox{1\columnwidth}{!}{
\setlength{\tabcolsep}{0.8mm}{
\centering
\begin{tabular}{lcccccc}
\toprule[1.5pt]
Removed Sentence & None & 1st  & 2nd  & 3rd  & 4th  & 5th  \\
\midrule
Object-Hal C-s ↓ & 51.0 & 49.8 & 49.6 & 47.0 & 39.4 & 37.0 \\ 
\bottomrule[1.5pt]
\end{tabular}
\centering
}}
\caption{Sentence-level hallucination rate of the response after removing sentences at difference positions.}
\label{tab:pos_fac}
\end{table}

\section{Quantitative Results}
\subsection{Empirical Evidence on Hallucination Observations}
\label{ss:c.halobsv}
The hallucination self-injection process in APASI is based on three key observations:
\textbf{object co-occurrence}, \textbf{language prior} and \textbf{positional factor}. These observations are not only consistent with prior studies but are also substantiated by our empirical analyses.
For \textbf{object co-occurrence}, Object-Hal evaluations of LLaVA-v1.5-7B’s responses reveal that $98\%$ of hallucinated objects co-occur with correct ones, with $25.9\%$ appearing as the top-1 co-occurring object and $67.9\%$ within the top 5.
For \textbf{language prior}, Fig.\ref{fig:pdmh} shows that models with fewer hallucinations yield higher PDM-H scores, indicating reduced reliance on language priors.
For \textbf{positional factor}, as shown in Tab.\ref{tab:pos_fac}, the hallucination rate decreases progressively as later sentences (from 1st to 5th) are removed from the response, indicating that hallucinations are more likely to appear in later parts of the response.
Together, these findings confirm the soundness of our design principles in capturing the key factors underlying hallucination generation.

\subsection{Performance Comparison with Larger Models}
To evaluate the practical competitiveness of APASI, we conduct comparisons against larger-scale models, specifically LLaVA-1.5-13B and LLaVA-1.6-13B. As shown in Tab.\ref{tab:larger}, despite the substantial parameter gap, our 7B models trained with APASI attain performance that is comparable to, or even surpasses that of the corresponding 13B models in the same series. These results further underscore the effectiveness and efficiency of our approach.

\begin{table}[t]
\resizebox{1\columnwidth}{!}{
\setlength{\tabcolsep}{0.8mm}{
\centering
\begin{tabular}{lcc}
\toprule[1.5pt]
Model                            &  Object-Hal C-s↓         & MMVet \\
\midrule
LLaVA-1.5-7B                     & 51.0                     & 30.5           \\
LLaVA-1.5-13B                    & 49.2                     & 33.5           \\
\textbf{LLaVA-1.5-7B+APASI-Base} & \textbf{38.1}            & \textbf{33.5}  \\
LLaVA-1.6-7B                     & 38.1                     & 42.5           \\
LLaVA-1.6-13B                    & 30.2                     & 43.9           \\
\textbf{LLaVA-1.6-7B+APASI-Base} & \textbf{28.8}            & \textbf{44.2}  \\
\bottomrule[1.5pt]
\end{tabular}
\centering
}}
\caption{Performance comparison with larger models.}
\label{tab:larger}
\end{table}

\begin{figure}[t]
	\begin{center}
		%\fbox{\rule{0pt}{2in} \rule{.9\linewidth}{0pt}}
		\includegraphics[width=\linewidth]{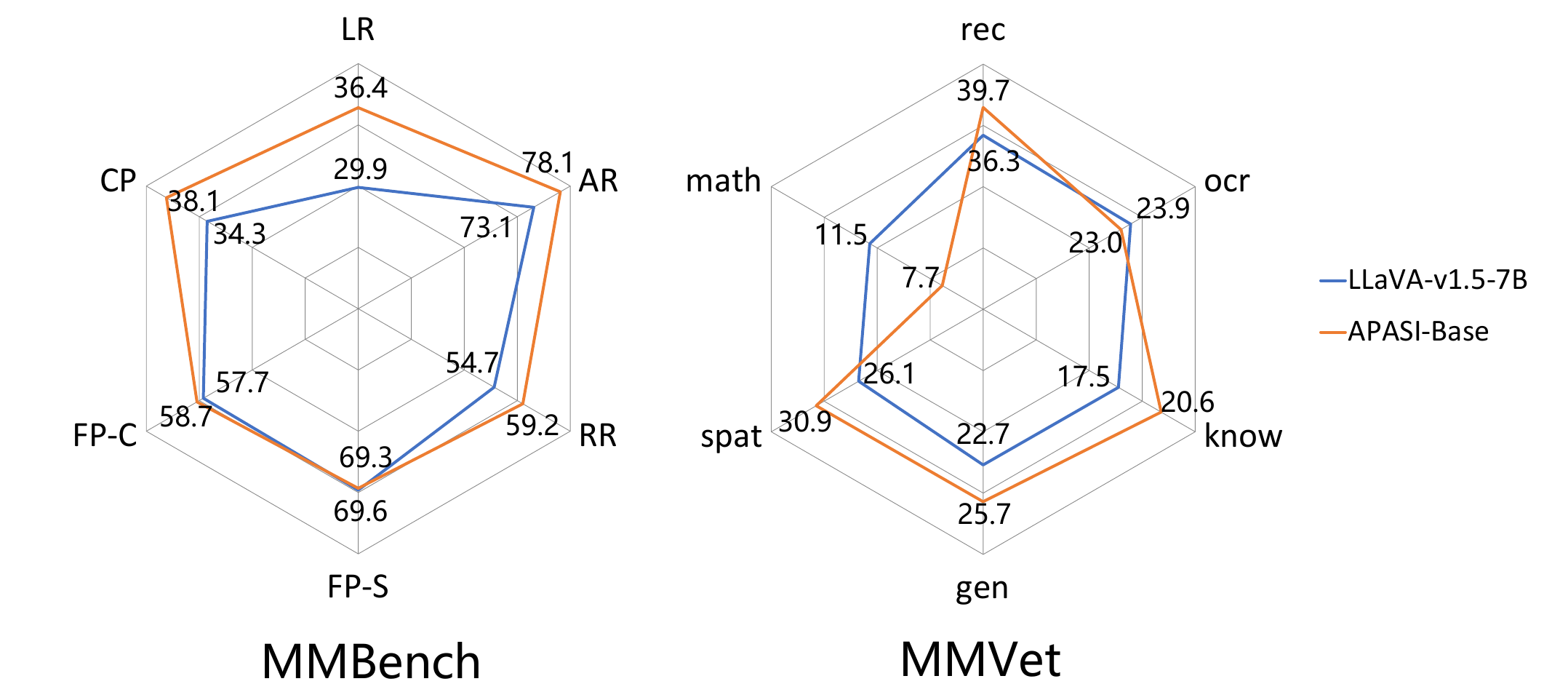}
	\end{center}
	\caption{Detailed Results on MMBench (Left) and MMVet (Right). Best viewed in color.}
	\label{fig:radar}
\end{figure}

\subsection{Detailed Results on Comprehensive Benchmarks}
% introduce the benchmarks
We evaluate comprehensive abilities of LVLMs on MMBench \cite{liu2025mmbench}, MMVet \cite{yu2023mm}, and LLaVABench \cite{liu2024visual}. 
MMBench evaluates the reasoning and perception capabilities and subdivides them into six Level-2 capability dimensions including: logic reasoning (LR), attribute reasoning (AR), relation reasoning (RR), fine-grained perception-single instance (FP-S), fine-grained perception-cross instance (FP-C), and coarse perception (CP).
MMVet evaluates six core capabilities of the LVLMs including recognition (rec), knowledge (know), OCR, Spatial awareness (Spat), Language generation (gen), and math.
LLaVA-BENCH evaluates the model's on three tasks including conversation (conv), detailed description (detail), and complex reasoning (complex).

Fig.\ref{fig:radar} shows detailed results on MMBench and MMVet. APASI-Base outperforms the baseline on all capabilities except math and OCR. This can be explained by the alignment target of hallucination mitigation. APASI focuses on improving the ability of perception for objects, but fails to cover perception for characters or mathematical symbols. Tab.\ref{tab:llavabench} shows detailed results on LLaVA-BENCH. APASI-Base outperforms the baseline on all three tasks. The improvement on the detailed description task is the highest among all. Notably, APASI achieves improvement on the reasoning capabilities on all three benchmarks, though the preference alignment in APASI doesn't optimize these capabilities directly. This suggests that reasoning abilities can benefit from the improvement of perception abilities.

\begin{table}[t]
\resizebox{1\columnwidth}{!}{
\setlength{\tabcolsep}{0.8mm}{
\centering
\begin{tabular}{lcccc}\toprule[1.5pt]
Model         & complex & conv   & detail & overall \\
\midrule
LLaVA-v1.5-7B & 79.5    & 51.8   & 56.6   & 65.5    \\
APASI-Base    & 80.5    & 53.1   & 60.7   & 67.3    \\
\midrule
$\Delta$              & 1.26\%  & 2.51\% & \textbf{7.24\%} & 2.74\%  \\
\bottomrule[1.5pt]
\end{tabular}
}}
\caption{Detailed Results on LLaVA-BENCH.}
\label{tab:llavabench}
\end{table}

\begin{table}[t]
\begin{center}
    \begin{tabular}{cccc}
        \toprule[1.5pt]
        \multirow{2}{*}{Synonym Sets} 
        & \multicolumn{2}{c}{Object-Hal} & AMBER \\
        \cmidrule(r){2-3} \cmidrule(r){4-4}
        & C-s $\downarrow$ & C-i $\downarrow$ & C-i $\downarrow$ \\ 
        \midrule
        WordNet(LVIS)         & 38.1           & 9.2         & 33.5 \\
        V3Det & \textbf{31.7} & \textbf{8.7} & \textbf{33.8} \\
        \bottomrule[1.5pt]
    \end{tabular}
\end{center}
\caption{Ablation studies on injection rates.}
\label{tab:symset}
\end{table}

\begin{table}[t]
\begin{center}
    \begin{tabular}{cccc}
        \toprule[1.5pt]
        \multirow{2}{*}{\Large $\rho$} 
        & \multicolumn{2}{c}{Object-Hal} & AMBER \\
        \cmidrule(r){2-3} \cmidrule(r){4-4}
        & C-s $\downarrow$ & C-i $\downarrow$ & C-i $\downarrow$ \\ 
        \midrule
        0.1         & 49.2           & 15.3         & 8.7 \\
        \textbf{0.2} & \textbf{38.1} & \textbf{9.2} & \textbf{6.0} \\
        0.3          & \textbf{38.1} & 10.5         & 6.1        \\
        0.4          & 42.4            & 11.2         & 6.2        \\
        0.5          & 41.3           & 10.9         & 6.2        \\
        0.6          & 46.4            & 12.3         & 6.4       \\
        \bottomrule[1.5pt]
    \end{tabular}
\end{center}
\caption{Ablation studies on injection rates.}
\label{tab:rate}
\end{table}

\subsection{Analysis on Synonym Sets}
The synonym set is applied during preprocessing to group object mentions when constructing the co-occurrence graph. This step offers lightweight, plug-and-play lexical normalization (e.g., merging “bicycle” and “bike”) to enhance the coverage of object relationships. To assess their impact, we replace the originally used WordNet (LVIS) \cite{miller1995wordnet,gupta2019lvis} with V3Det \cite{wang2023v3det}, a more modern synonym set featuring broader and finer-grained categories. As shown in Tab.\ref{tab:symset}, this substitution yields improvements of +6.4/+0.5/+0.3 on ObjectHal-C-s/ObjectHal-C-i/MMVet, respectively. These results highlight the flexibility and extensibility of our pipeline with respect to synonym sources.

\subsection{Ablation Studies on Injection Rate}
\label{ss:b.2}
We design ablative experiments on the injection rate $\rho$.
As shown in Tab.\ref{tab:rate}, APASI achieves the best performance when setting $\rho$ to 0.2. When $\rho$ is greater than 0.2, APASI performs poorer as $\rho$ increases. This is consistent with the observation that larger gap within the preference pair makes the alignment task more easy, which is less effective for model improvement. Setting $\rho$ to 0.1 results in worse performance than 0.2. An injection of 0.1 leads to a situation where number of replaced sentences is zero in about half of the preferred sentences, thus making the corresponding pair invalid.

\begin{table}[t]
\begin{center}
    \resizebox{0.98\columnwidth}{!}{
    \setlength{\tabcolsep}{1mm}{
\begin{tabular}{lll}
\toprule[1.5pt]
Method                & Data Source                  & Size        \\
\midrule
POVID                 & LLaVA-Instruct               & 17k         \\
HA-DPO                & VG                           & 6k \\
RLAIF-V               & COCO, MovieNet,... (7 total) & 83k         \\
CLIP-DPO              & COCO, SAM,... (12 total)     & 750k        \\
CSR                   & LLaVA-Instruct               & 13k         \\
STIC                  & COCO, LLaVA-Instruct         & 11k         \\
SIMA                  & LLaVA-Instruct               & 17k         \\
\midrule
Ours SI-23k  & LLaVA-Instruct               & 23k         \\
Ours SI-130k & LLaVA-Instruct, VG           & 130k        \\
\bottomrule[1.5pt]
\end{tabular}
    }
    }
\end{center}
\caption{Comparison of preference data sources and sizes.}
\label{tab:datasize}
\end{table}

\begin{table}[t]
\begin{center}
    \resizebox{0.9\columnwidth}{!}{
    \setlength{\tabcolsep}{1.5mm}{
    \begin{tabular}{lccc}
        \toprule[1.5pt]
        \multirow{2}{*}{Preference Data} 
        & \multicolumn{2}{c}{Object-Hal} 
        & \multirow{2}{*}{MMVet} \\
        \cmidrule(r){2-3}
        & C-s $\downarrow$ & C-i $\downarrow$ & \\ 
        \midrule
        LLaVA-v1.5-7B	& 51.0 & 13.7 & 30.5 \\
        SI-23k (APASI-Base) & 38.1 & 9.2 & 33.5  \\ 
        \midrule
        SI-6k & 46.0 & 12.9 & 31.7 \\
        SI-23k + POVID-17k   & \textbf{34.9} & \textbf{8.3} & \textbf{34.1}    \\
        \bottomrule[1.5pt]
    \end{tabular}
    }
    }
\end{center}
\caption{Performance of APASI trained on different preference data.}
\label{tab:data}
\end{table}

\subsection{Analysis on Data Source}
To enable a fair comparison with existing methods, in Tab.~\ref{tab:datasize} we provide a detailed analysis of data sources and dataset sizes across all methods compared. Our SI-23k and SI-130k datasets are constructed from the simplest sources, relying solely on the supervised fine-tuning (SFT) dataset of the baseline model (LLaVA-v1.5). Notably, APASI-Base and APASI-IACL, trained on the moderate-scale SI-23k dataset, already achieve competitive performance.

To further evaluate performance under limited data conditions, we trained an APASI variant using only 6k samples randomly drawn from SI-23k. As shown in Tab.~\ref{tab:data}, APASI with 6k data still outperforms the baseline by +5.0/+0.8/+1.2 on ObjectHal-C-s/ObjectHal-C-s/MMVet, respectively. The performance gap between the 6k and 23k settings underscores the critical role of data scale in enhancing model effectiveness.

We also extend the data sources by incorporating GPT-4V-labeled preference pairs, combining SI-23k with POVID-17k~\cite{zhou2024aligning}. Tab.~\ref{tab:data} reports the results of APASI trained on SI-23k alone versus the combined dataset.
Incorporating the external dataset yields modest improvements of +3.2/+0.9/+0.6 on ObjectHal-C-s/ObjectHal-C-i/MMVet, respectively. Although APASI is designed to function without reliance on external resources, these results indicate that it can also be effectively scaled when such resources are available.

% \textbf{1)} We combine the self-injection dataset SI-23k with POVID-17k which contains GPT-4V-labeled preference pairs. This combined training yields a modest improvement of \textbf{+3.2/+1.0} on ObjectHal-C-i/MMVet, compared to APASI-Base trained with SI-23k, suggesting the potential of the combined strategy.
% \textbf{2)} We'd like to emphasize that the main strength of APASI lies in its independence from external resources, but it can also be scaled with external data when available. We'll add this to the final version.

\subsection{Computational Cost of APASI}
We analyze the computational cost of APASI using the SI-23k dataset and LLaVA-v1.5-7B as the target LVLM. 

Preference data construction includes the following stages: 
\textbf{1) Preferred generation.} The target LVLM generates preferred responses with input images and textual questions taking about 80 mins on 8 V100 32GB GPUs.
\textbf{2) Co-occurrence graph construction.} Objects in preferred responses are parsed via WordNet toolbox and synonym sets. We traverse these objects to build the co-occurrence graph stored as a dictionary, where each key-value pair represents an object and its co-occurring objects with frequencies, requiring about 6MB storage. This takes about 3 mins with one Intel E5-2698 CPU process.
\textbf{3) Hallucination injection.} The visually disabled target LVLM generates hallucinated sentences guided by co-occurring objects, replacing the sampled sentences in the preferred responses. This takes about 80 mins on 8 V100 32GB GPUs, with the resulting dataset occupying about 8MB. 

As training takes about 360 mins, data construction accounts for only about $31.2\%$ ($\frac{80+3+80}{80+3+80+360}$) in a full iteration's time. Storage and hardware demands also remain acceptable, making the data construction practically feasible. 

\begin{figure*}[t]
	\begin{center}
		%\fbox{\rule{0pt}{2in} \rule{.9\linewidth}{0pt}}
		\includegraphics[width=0.9\linewidth]{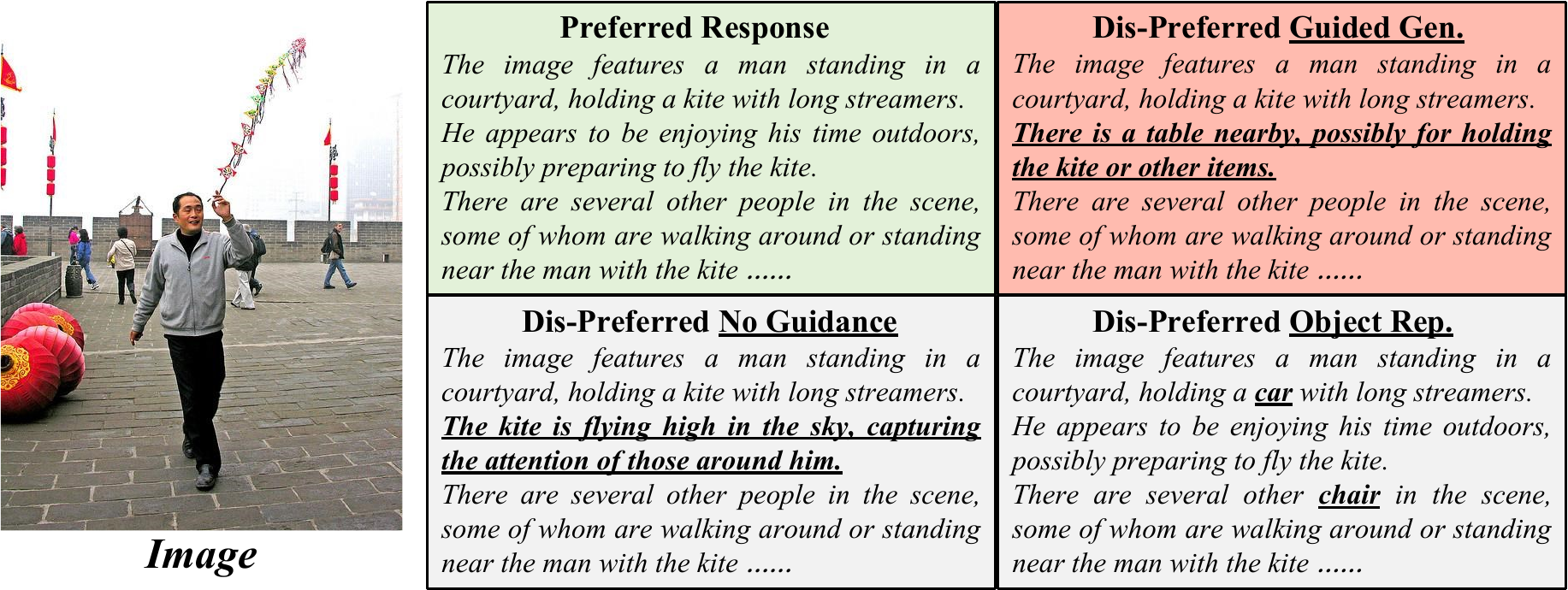}
	\end{center}
	\caption{
    Comparison of the dis-preferred responses under different hallucination injection settings. 
    The preferred response is in the light green box. 
    The dis-preferred response obtained by guided model generation, which is actual used in APASI, is in the light red box. 
    The two ineffective dis-preferred responses  are in the gray boxes.
    The injected hallucination parts are {\ul underlined}. Best viewed in color.}
	\label{fig:disp}
\end{figure*}

\section{Qualitative Results}
\subsection{Analysis on Injection Settings}
\label{ss:c.1}
We compare dis-preferred responses under different hallucination injection settings, as shown in Fig.\ref{fig:disp}. For hallucination injection in APASI, the target model is guided to generate sentences about non-existent {\ul table} to obtain the dis-preferred responses. 
In this way, the injected sentence is ensured is guaranteed to contain hallucinations, and the injected hallucination remains linguistically reasonable.
If removing the guidance of the co-occurring object, the model generates an actually correct sentence describing the kite for injection. The dis-preferred response without hallucinations makes the preference pair invalid.
If injecting hallucinations only by replacing with the word {\ul car} and {\ul chair}, the dis-preferred response includes absurd mistakes in the like ``man holding cars''. These absurd mistakes are inconsistent with typical hallucinations with linguistic reasonability, thereby the  preference pair less effective.

\subsection{Examples of Model Response}
\begin{figure*}[t]
	\begin{center}
		%\fbox{\rule{0pt}{2in} \rule{.9\linewidth}{0pt}}
		\includegraphics[width=\linewidth]{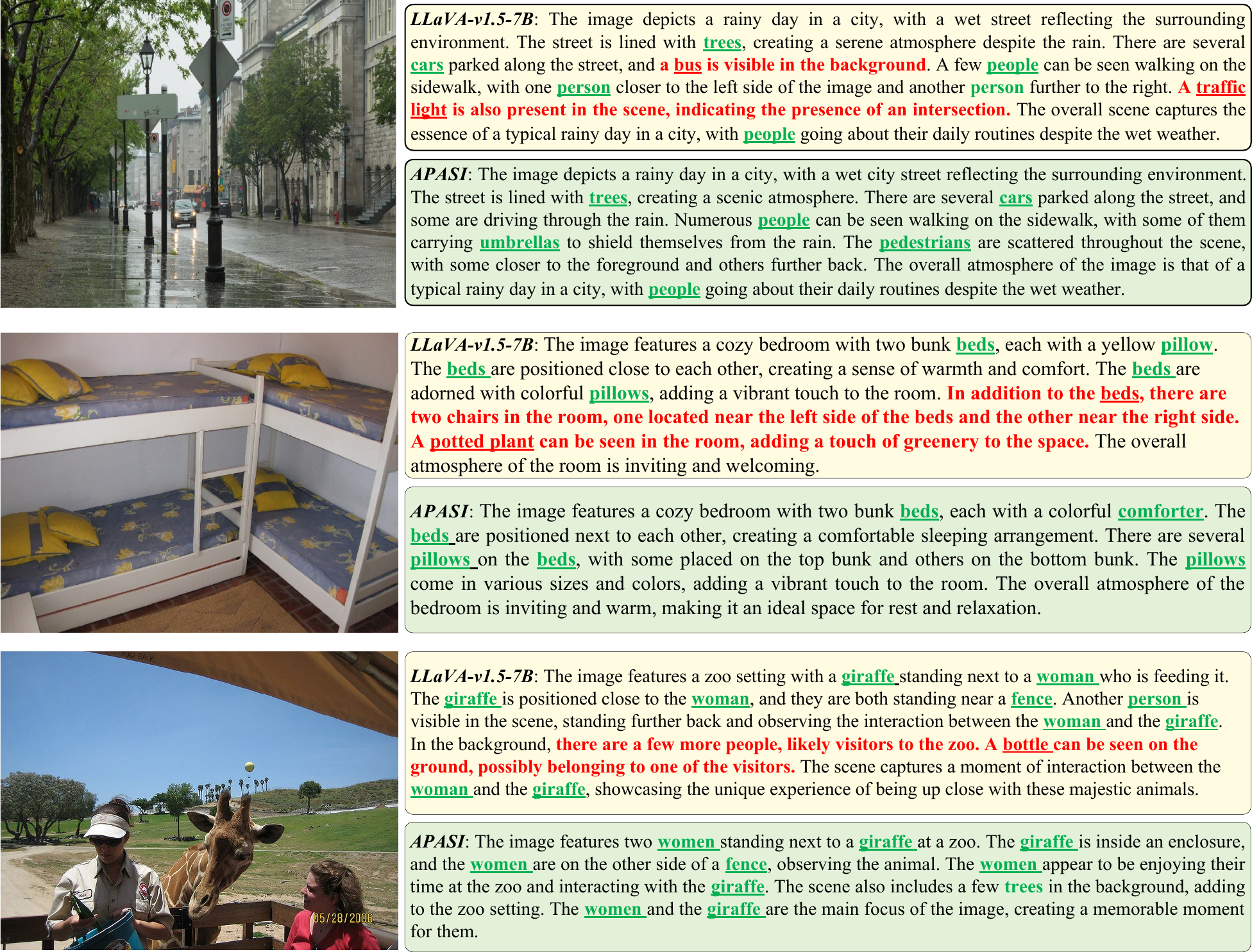}
	\end{center}
	\caption{Comparison of the responses of the baseline and APASI. Correct and wrong objects are underlined in red and green respectively. Other hallucinated contents are in red. Best viewed in color.}
	\label{fig:case1}
\end{figure*}

\begin{figure*}[t]
	\begin{center}
		%\fbox{\rule{0pt}{2in} \rule{.9\linewidth}{0pt}}
		\includegraphics[width=\linewidth]{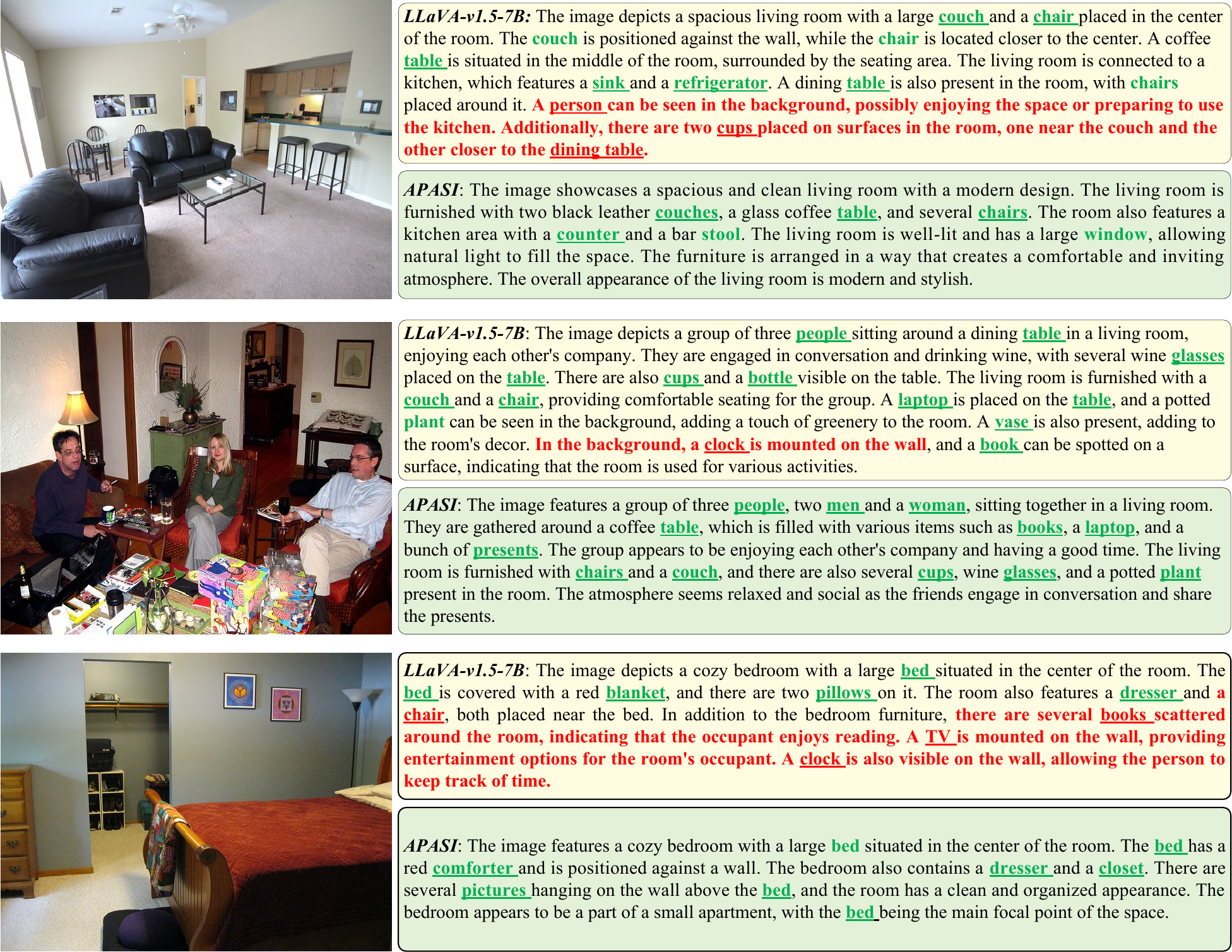}
	\end{center}
	\caption{Comparison of the responses of the baseline and APASI. Correct and wrong objects are underlined in red and green respectively. Other hallucinated contents are in red. Best viewed in color.}
	\label{fig:case2}
\end{figure*}

Fig.\ref{fig:case1} and Fig.\ref{fig:case2} show more comparative examples of APASI and the LLaVA-v1.5-7B baseline. Both models are given with descriptive prompts and images from COCO-test2017 \cite{lin2014microsoft} or MMHal \cite{sun2023aligning}. Results show that APASI effectively mitigates the hallucination problem for LLaVA-v1.5-7B while capturing all major objects in the image.

\end{document}